%% file: main.tex
\documentclass[twoside,accepted]{article}
\usepackage{template/aistats2024}

\usepackage{template/macro}
\input{template/package}

\renewcommand\paragraph[1]{\noindent\textbf{#1}}

\date{\today}

\begin{document}

\twocolumn[
\aistatstitle{The Galerkin method beats Graph-Based Approaches for Spectral Algorithms}
\aistatsauthor{Vivien Cabannes \And Francis Bach}
\aistatsaddress{ Meta AI \And  INRIA, Ecole Normale Supérieure} ]

\begin{abstract}
    \input{abstract}
\end{abstract}

\input{core}

\paragraph{Acknowledgement.}
The authors would like to thank Loucas Pillaud-Vivien and Ricky Chen for fruitful discussions.

\bibliography{main}

\appendix
\onecolumn

\input{appendix/proof}

\input{appendix/experiments}

\end{document}

%% file: template/package.tex
\usepackage[T1]{fontenc}
\usepackage[utf8]{inputenc}
\usepackage[english]{babel}

\usepackage{times}

\PassOptionsToPackage{hyphens}{url}
\usepackage{url}
\usepackage{hyperref}

\usepackage{graphicx}
\usepackage{caption}
\usepackage{tikz}
\usepackage{pgfplots}
\pgfplotsset{compat=1.17}

\usepackage{array}
\usepackage{booktabs}
\usepackage{microtype}
\usepackage{xcolor}
\usepackage{enumitem}
\setitemize{itemsep=0pt,parsep=0pt,topsep=0pt}
\setenumerate{itemsep=0pt,parsep=0pt,topsep=0pt}
\usepackage[ruled,vlined]{algorithm2e}

\usepackage[sectionbib,authoryear,round]{natbib}
\usepackage[normalem]{ulem}

%% file: abstract.tex
Historically, the machine learning community has derived spectral decompositions from graph-based approaches. We break with this approach and prove the statistical and computational superiority of the Galerkin method, which consists in restricting the study to a small set of test functions. In particular, we introduce implementation tricks to deal with differential operators in large dimensions with structured kernels. Finally, we extend on the core principles beyond our approach to apply them to non-linear spaces of functions, such as the ones parameterized by deep neural networks, through loss-based optimization procedures.

%% file: core.tex
\section{INTRODUCTION}

Eigen and singular decompositions are ubiquitous in applied mathematics.
They can serve as a basis to define good features in machine learning pipelines \citep{Belkin2003,Coifman2006,Balestriero2022}, while a set of good features naturally define pullback distances on the original data.
Those features and distances are naturally referred to as ``spectral embeddings'' and ``spectral distances''.
The latter are thought to provide meaningful geometries on the data, which explain their uses for clustering \citep{Belkin2004,Schubert2018DBSCAN}, as well as diffusion models \citep{chen2023riemannian}.
In the machine learning community, spectral decompositions are usually derived from the eigen decompositions of different graph Laplacians built on top of the data \citep{Chung1997,Zhu2003,Ham2004}.
However, those methods are known to scale poorly with the input dimension \citep{Bengio2006,Singer2006,Hein2007}, although they had applications in many different fields, such as molecular simulation \citep{Glielmo2021}, acoustics \citep{Bianco2019} or the study of gene interaction \citep{VanDijk2018}.

In this paper, we suggest a different approach to approximate the spectral decompositions of a large class of operators.
Our method consists in restricting the study of infinite-dimensional operators on a basis of simple functions, which is usually referred to as Galerkin, Ritz or Raleigh methods \citep{Singer1962}, if not Bubnov or Petrov \citep{Fluid}, depending on the research community.
We make the following contributions.
\begin{enumerate}
    \item
    We release an algorithm to compute spectral decompositions of a large class of operators that is {\em statistically and computationally efficient},\footnote{%
        Our Python library can be downloaded through the command line \texttt{\$ pip install klap}, or built from the \href{https://github.com/VivienCabannes/laplacian}{source code}.
    }
    and provide experiments to confirm those theoretical findings with numerical analysis.
    \item 
    We show that our method prevails over graph-based approaches both statistically and computationally.
    This opens exciting follow-ups for any spectral-based algorithms, could it be spectral clustering, spectral embeddings, or spectral distances.
    \item 
    We show that our method prevails over kernelized algorithms based on the use of representer theorems \citep{Scholkopf2001representer,Zhou2008}. 
    Interestingly, our Galerkin approach can be seen as unifying and generalizing both random features and the Nystrom method.
    To take a concrete example, for the Laplacian problem considered by \citet{PillaudVivien2023}, their use of the representer theorems leads to implementation requiring $O(n^3d^3)$ flops, while the instantiation of our generic framework provides an implementation in $O(n^2 + n^{3/2}d)$ flops without any loss from a statistical viewpoint, for $n$ the number of samples, and $d$ the sample dimension.
    This opens exciting follow-ups for any kernel methods dealing with derivatives, such as Hermite regression.
    \item
    Finally, we extract the core principles behind our method in order to consider non-linear spaces of functions through loss-based optimization procedures. 
    We equally discuss how our perspective can model approaches that have led to state-of-the-art results in self-supervised learning.
\end{enumerate}

\section{SETUP}

This section details the setup, motivations, and a running example behind our study.

\paragraph{Data.}
We assume that $n$ samples $(x_i)_{i\in[n]}$ have been collected and stored as raw vectors $x_i \in \X = \R^d$.\footnote{%
    In all the following, we use the notation $[n]=\{1, 2, \ldots, n\}$.
}
The collection process is idealized as underlying a distribution $\rho \in \prob{\X}$ which has generated the samples as $n$ independent realizations of the variable $X\sim\rho$.

\paragraph{Goal.}
Let us consider an operator $\cL:L^2(\rho)\to L^2(\rho)$ in a large class of operators.
To be precise, we assume that $\cL$ has discrete spectrum and defines a bilinear form as an expectation over the data through a known operation $H:L^2(\rho)\times L^2(\rho)\times\X\to\R$ that is supposed to be bilinear in $f$ and $g$,
\begin{equation}
    \label{eq:energy}
    \cE_{\cL}(f, g) = \scap{f}{\cL g}_{L^2(\rho)} = \E_X[H(f,g,X)].
\end{equation}
For example, $\cL$ could be equal to the differential operator defined with the partial derivative $\partial_i$, in which case $H(f, g, x) = f(x)\partial_i g(x)$.
Our goal is to approximate the spectral decomposition of $\cL$
\begin{equation}
  \label{eq:spe-dec}
  \cL = \sum_{i\in\N^*} \lambda_i f_i \otimes g_i,
\end{equation}
for $\lambda_i > 0$ the ordered singular values of $\cL$ and $f_i, g_i \in L^2(\rho)$ the corresponding left and right singular functions.
The decomposition is performed in $L^2(\rho)$, i.e., the $(f_i)$, and respectively the $(g_i)$, are orthonormal in $L^2(\rho)$.

\paragraph{Running example.}
A prototypical example is provided by 
\begin{equation}
    \label{eq:lap}
    H_0(f, g, X) = \scap{\nabla f(X)}{\nabla g(X)}.
\end{equation}
The resulting operator is the Laplacian $\cL_0 = \nabla^* \nabla$ where $\nabla$ is the Euclidean gradient, and the adjoint is taken with respect to the $L^2(\rho)$-geometry.\footnote{%
    Laplacians are usually defined as negative self-adjoint operators $-\nabla^*\nabla$ as for the usual Laplacian $\Delta = \sum \partial_{ii}^2$ which corresponds to adjunction with respect to $L^2(\diff x)$ endowed with the Lebesgue measure.
    This paper rather uses the graph-based convention where Laplacians are positive self-adjoint operators.
}
In this case, $\cL_0$ is positive self-adjoint, hence $f_i = g_i$.

\begin{figure}[t]
    \centering
	\includegraphics[width=.4\textwidth]{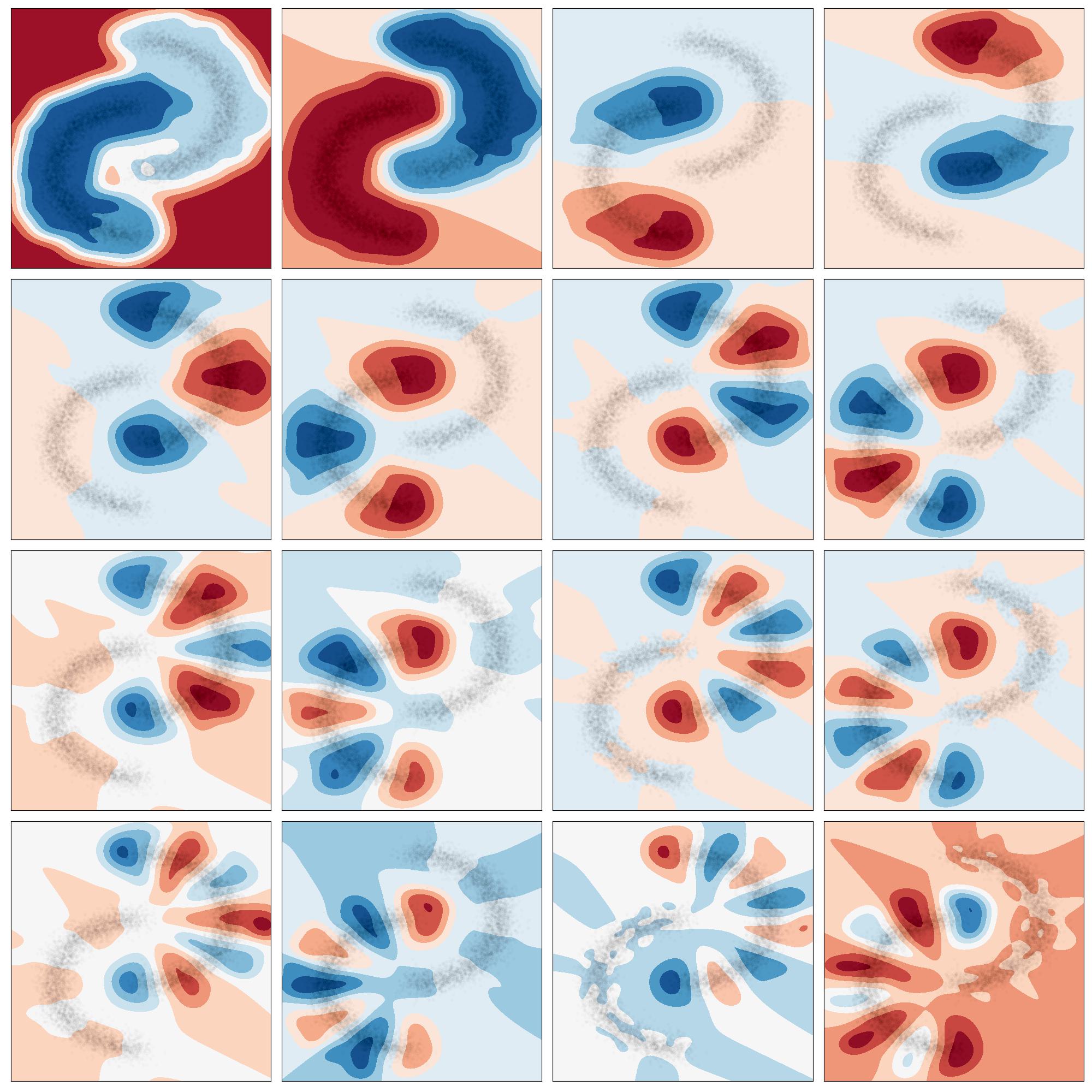}
	\caption{
	  Level lines of the first sixteen learned eigenfunctions of $\cL_0$ when the data generates two half-moons with $d=2$, with Algorithm~\ref{alg:klap}, $n=10^5$ points, and $p=200$ Galerkin functions derived from the exponential kernel.
	  See how those eigenfunctions are separated between the two clusters, and how, on each cluster, they identify with Fourier modes (i.e., cosines) when distorting the segment $[0,1]$ into a half-moon.
	}
    \label{fig:Fourier}
\end{figure}

The operator $\cL_0$ has found applications for representation learning, that aims to extract good feature to describe raw data, as the eigenfunctions $f_i$ encode some sort of ``modes'' on the input space, while the $\lambda_i$ captures a notion of complexity of those \citep{Bousquet2003,Cabannes2023a}, similarly to Fourier modes where complexity increases with frequency, see Figure~\ref{fig:Fourier}.

Interestingly, $\cL_0$ also relates to Langevin dynamics, which aims to generate new samples that could have originated from the original data distribution \citep{Dockhorn2022}.
In particular, when $\rho$ derives from a potential $V:\X\to\R$, i.e., $\rho(\diff x) = \exp(-V(x))\diff x$ with $\diff x$ the Lebesgue measure, under mild assumptions on $V$, as detailed in Appendix~\ref{app:char-L0}, the operator $\cL_0$ is characterized by
\[
    \cL_0:f\mapsto -\Delta f + \scap{\nabla V}{\nabla f}.
\]
In this setting,~\eqref{eq:energy} is known as the Dirichlet energy, and $\cL_0$ as the infinitesimal generator of the Langevin diffusion \citep{Grenander1994,Bakry2014,Liu2016}.\footnote{%
    Reciprocally, diffusion models have been proposed as some form of power methods to estimate eigen decomposition by \citet{Han2020}.
}

\section{GALERKIN METHOD}
This section introduces our method, and discusses its statistical efficiency.
Our method assimilates to the Galerkin method, together with a Monte-Carlo quadrature rule to estimate the scalar product of $L^2(\rho)$ \citep{Chatelin1983}.

\begin{algorithm}[ht]
    \KwData{Data $(x_i) \in \X^n$, function $H$ as per~\eqref{eq:energy}.}
    Choose $p$ test functions $(\phi_i)_{i\in[p]}$, $(\psi_i)_{i\in[p]}$;\\
    Compute $\hat L = (\sum_{k\in[n]} H(\phi_i, \psi_j, x_k))_{ij} \in \R^{p\times p}$;\\ 
    Compute $\hat\Phi = (\sum_{k\in[n]} \phi_i(x_k) \phi_j(x_k))_{ij} \in \R^{p\times p}$;\\
    Compute $\hat\Psi = (\sum_{k\in[n]} \psi_i(x_k) \psi_j(x_k))_{ij} \in \R^{p\times p}$;\\
    Solve $(\Lambda,A,B) = \operatorname{GSVD}(\hat L, \hat\Phi,\hat\Psi)$~\eqref{eq:gsvd};\\
    Set $\hat \lambda_i := \Lambda_{ii}$, $\hat f_i := \sum_{j\in[p]} A_{ij} \phi_i$, $\hat g_i := \sum B_{ij} \psi_i$;\\
    \KwResult{Estimate $(\hat \lambda_i, \hat f_i, \hat g_i)$ to decompose $\cL$ as per~\eqref{eq:spe-dec}.}
    \caption{Galerkin method}
    \label{alg:galerkin}
\end{algorithm}

\subsection{Algorithm}
Rather than working with the infinite-dimensional operator~$\cL$, we will restrict our study to its action on a finite-dimensional space of functions.
To do so, we consider $p$ real-valued functions $(\phi_i)_{i \in [p]}$ to cast a vector $\alpha\in\R^p$ into a function in $L^2(\rho)$ through the embedding
\begin{equation}
    \begin{array}{cccc}
         F:&\R^p &\to& L^2(\rho);  \\
         &\alpha &\mapsto&\sum_{i\in[p]}\alpha_i \phi_i, 
    \end{array} 
\end{equation}
and we search for the singular functions as $f_i = F \alpha_i$ for $\alpha_i\in\R^p$.
In full generality, we can distinguish the search space for left and right singular functions, and introduce a second family of real-valued functions $(\psi_i)_{i\in[p]}$ to map $\beta\in\R^p$ to a function in $L^2(\rho)$ through the embedding $G$ defined from the $(\psi_i)$.
Restricted to those parametric functions, the operator $\cL$ assimilates to the {\em matrix} $L \in \R^{p \times p}$, defined as
\begin{equation}
    \label{eq:L}
    L_{ij} := (F^*\cL G)_{ij} = (\E_X[H(\phi_i, \psi_j, X)])_{ij},
\end{equation}
where the adjunction is taken with respect to $L^2(\rho)$ and $\R^p$, and the last equation is proven in Appendix~\ref{app:operator}.

The search for the decomposition~\eqref{eq:spe-dec} reduces to the search of the decomposition,
\[
    L = \sum_{i\in\N^*} \lambda_i F^* f_i g_i^* G
    \approx \sum_{i\in[t]} \lambda_i F^* F \alpha_i \beta_i^* G^* G,
\]
where the last equation is due to our search for $f_i$ and $g_i$ under the form $F\alpha_i$ and $G\beta_i$, together with a threshold $t\in\N$.
Introducing the $t\times p$ matrices $A = (\alpha_i)_{i\in[t]}$, $B = (\beta_i)$ and $\Lambda = \diag(\lambda_i) \in \R^{t\times t}$, the last equation is rewritten as
\[
    L \approx F^*F A^\top \Lambda B G^* G.
\]
In addition to the above decomposition, we should specify orthogonality constraints,
\[
    \scap{F\alpha_i}{F\alpha_j}_{L^2(\rho)} = \alpha_i F^* F \alpha_j = \alpha_i \Phi \alpha_j = \delta_{ij},
\]
where $\Phi \in \R^{p\times p}$ is defined as (see Appendix~\ref{app:operator})
\begin{equation}
    \label{eq:Phi}
    \Phi_{ij} = (F^* F)_{ij} = (\E[\phi_i(X) \phi_j(X)])_{ij}.
\end{equation}
Considering empirical averages in the definition of $L$ in~\eqref{eq:L}, $\Phi$ and $\Psi$ in~\eqref{eq:Phi} leads to Algorithm~\ref{alg:galerkin}, where the generalized singular value decomposition (GSVD) of $(\hat L, \hat\Phi, \hat\Psi)$ reads
\begin{equation}
    \label{eq:gsvd}
    A \hat L B^\top = \Lambda,\qquad
    A \hat\Phi A^\top = B\hat\Psi B^\top = I,
\end{equation}
where $A, B, \Lambda\in\R^{p\times p}$ and $\Lambda$ is diagonal with positive entries.
This decomposition exists as soon as $t = p$ and $\hat \Phi$ and $\hat \Psi$ are invertible, as shown in Appendix~\ref{app:gsvd}.

\paragraph{Computational complexity.}
Our method consists in building and storing the $p\times p$ matrices $\hat L$, $\hat\Phi$ and $\hat\Psi$, before solving the associated generalized singular value decompositions.
The building of the matrix $\hat L$ scales in $O(np^2 c_H)$ flops, where $c_H$ is the cost of evaluating $H(\phi_i, \phi_j, x_k)$, while the decomposition scales in $O(p^3)$.
As a consequence, the overall number of flops is $O(np^2c_H + p^3)$.
In terms of memory, we only need $O(p^2 + b_H)$ bits of memory where $b_H$ is the memory needed to evaluate a single $H(\phi_i, \psi_j, x_k)$.

\subsection{Statistical Efficiency}

Our algorithm can be seen through two actions.
First, the operator $\cL$ in~\eqref{eq:energy} is restricted to its action on $\ima F$ and~$\ima G$ with the introduction of the operator $L$ in~\eqref{eq:L}.
Second, because $\ima L$ and $\ima G$ are finite-dimensional, the operator $L$ assimilates to a matrix.
This matrix is defined from the distribution $\rho$ but is approximated with empirical data, leading to $\hat L$, which will concentrate around $L$ as the number of data grows.
Those facts are captured formally by Theorem~\ref{thm:stat-phi}, proven in Appendix~\ref{app:stat-phi}.
It introduces $\Pi_F$ the orthogonal projector on $\ima F$ in $L^2(\rho)$.
It focuses on the reconstruction of the inverse $\cL^{-1}$ in operator norm, based on the estimation suggested by Algorithm~\ref{alg:galerkin}.
Controlling the operator norm allows the reconstruction of both the spectral values and functions \citep[e.g.,][]{Weyl1912}.
Our choice to focus on $\cL^{-1}$ is due to the fact that when $\cL$ is a differential operator, $\cL$ is usually not bounded, but its inverse is compact \citep{Kondrachov1945certain}.

\begin{theorem}
	\label{thm:stat-phi}
	Assume that $H(\phi_i, \psi_j, x)$ is bounded by $H_\infty$ independently of $(i, j, x)$, and that $L$ is invertible.
	For any $\delta > 0$, and $n > 3 \max(1, p^2 H_\infty^2 \|L^{-1}\|^{-2}) \log(2p/\delta)$, the following holds true with probability at least $1-\delta$ (the randomness coming from the data),
    \begin{align}
		&\!\!\!\!\big\| \cL^{-1} - G \hat L^{-1} F^* \big\|
        \nonumber
		\\[-.2cm]&\qquad\leq  3 \|\Psi^{1/2} L^{-1}\| \|L^{-1}\Phi^{1/2}\|\sqrt{\frac{8p^2H_\infty^2}{3n}\log(\frac{2p}{\delta})}.
        \nonumber
		\\&\qquad\qquad+ \norm{(I-\Pi_G)\cL^{-1}}   
		+ \norm{\cL^{-1}(I-\Pi_F)},
    \end{align}
    where $\norm{\cdot}$ is the operator norm.
\end{theorem}

The first term in Theorem~\ref{thm:stat-phi} can be estimated empirically, it relates to the variance of our estimate. 
The second and third term relates to the bias of our estimator, and cannot be known without specific assumptions on the spectral decomposition of the operator $\cL$.

In the following, we will consider that the features $\phi_i$ and $\psi_i$ were actually obtained as independent realizations of a random variable $\phi$ and $\psi$.
In those settings, we introduce the following operators in $L^2(\rho)$
\[
    \Sigma = \E_\phi[\phi\phi^*], \qquad \Xi = \E_\psi[\psi\psi^*],
\]
where the adjunction is understood in $L^2(\rho)$.
The projection on $F$ naturally becomes the projection on $\ima\Sigma^{1/2}$.
The following theorem, whose proof is provided in Appendix~\ref{app:stat}, refines Theorem~\ref{thm:stat-phi} in this particular situation.

\begin{theorem}
    \label{thm:stat}
        Let $(\psi_i)_{i \in [p]}$ be $p$ independent realizations of a random function $\psi$.
	Let $\Pi_p$ denotes the projection on the span of the first $p$ spectral functions of $\Xi$, and $\cL_p = \sum_{i\in[p]} \lambda_i f_ig_i^*$.
	Assume that $\|\psi\|_{L^2(\rho)} \leq M$ almost surely.
	In the setting of Theorem~\ref{thm:stat-phi}, for any~$\delta$, with probability $1-\delta$ (the randomness being understood with respect to the random functions), when $p > 3\log(2/\norm{\Sigma}\delta)$,
	\begin{align}
	  &\!\!\!\norm{(I-\Pi_G)\cL^{-1}}
	  \leq \lambda_{p+1}^{-1} + \norm{(I-\Pi_p)\cL_p^{-1}} 
	  \nonumber
	  \\&\qquad\qquad\qquad + \norm{\Pi_p\Sigma^{-1}\cL_p^{-1}}\sqrt{\frac{8M^2}{3p}\log(2p/\delta)}.
	\end{align}
\end{theorem}

In Theorem~\ref{thm:stat}, $\lambda_{p+1}^{-1}$ captures how fast the spectrum of $\cL^{-1}$ vanishes, $\norm{(I-\Pi_p)\cL_p^{-1}}$ depends on how well-specified our model is to retrieve the first $p$ spectral functions of $\cL$, and $\norm{\Pi_p\Xi^{-1}\cL_p^{-1}}$ relates to the variance of our estimator.
In order to understand the operator norms appearing in the previous theorems, notice that when $f_i$ and $g_i$ are singular functions of $\Sigma$ and $\Xi$, we have
\[
    \|\Sigma^{-1/2} \cL^{-1}\Xi^{-1/2}\| = \sup_{i\in\N^*} \lambda_i^{-1} \|\Sigma^{-1/2}f_i\| \cdot \|\Xi^{-1/2}g_i\|.
\]
In other terms, ensuring this norm to be bounded consists in enforcing that the complexity to reconstruct $f_i$ from $\phi$ (captured with $\|\Sigma^{-1/2}f_i\|$) and $g_i$ from $\psi$ does not grow too fast compared to the vanishing rates of the sequence~$(\lambda_i)$.
In particular, when $(\lambda_i)$ decreases slowly, we need to ensure $\|\Sigma^{-1/2}f_i\|$ to be small for many indices in order to guarantee good reconstruction properties, while when it decreases fast, only approximating well the first few spectral functions guarantee a similar overall reconstruction error.
Those considerations are illustrated with a concrete example in Section~\ref{sec:lap}.

\begin{figure*}[t]
    \centering
    \includegraphics{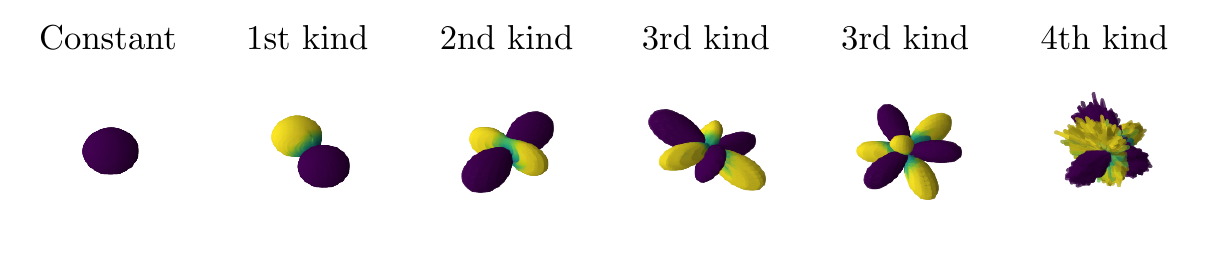}
    \includegraphics{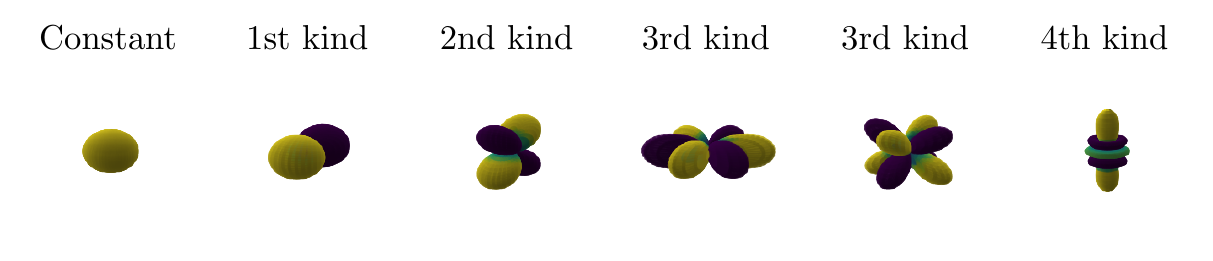}
    \vspace{-2em}
    \caption{
      Learning spherical harmonics with polynomials of degree three (with $k_x(y) = (1 + x^\top y)^3$ which corresponds to features maps that concatenates all the multivariate monomials of degree smaller or equal to $s=3$).
      Because we consider $\rho$ uniform on the sphere, the operator $\cL_0$ is diagonalized by spherical harmonics, which are polynomials of increasing degrees.
      The polynomial kernel of degree $D$ allows to learn all harmonics of $s$-th kind for $s$ smaller or equal to $D$ (the ones of higher kind are polynomials of higher degree that can not be reconstructed with polynomials of degree $D$ as illustrated with the fourth kind on the figure).
      Some of the learned eigenfunctions are represented on the top row, while some ground truths are represented on the bottom row.
      Our method learned perfectly valid harmonics, although, for eigenvalues that are repeated, it does not learn the canonical ones, but any basis of the different eigenspaces (which can be observed with the harmonics of the second kind in the figure).
    }
    \label{fig:spherical}
\end{figure*}

\section{DIFFERENTIAL OPERATORS, INVARIANT KERNELS}
This section discusses a large class of operators that take the form of equation~\eqref{eq:energy}, and natural spaces of functions to instantiate our algorithm.

\paragraph{Differential operators.}
A large class of interest lies in the operators defined through the bilinear form~\eqref{eq:energy} with
\begin{equation}
    \label{eq:differential}
    H_c(f, g, x) = \sum_{\alpha,\beta \in \N^d} c_{\alpha,\beta} \partial_\alpha f(x) \partial_\beta g(x),
\end{equation}
where for $\alpha = (\alpha_1, \dots, \alpha_d) \in \N^d$, $\partial_\alpha f$ denotes the partial derivatives of $f$ where the $i$-th coordinate is derived $\alpha_i$-times, and $c = (c_{\alpha, \beta})$ is a sequence in $\R$ indexed by $\N^d\times\N^d$ with a finite number of non zero elements.
This class encompasses the operator $\cL_0$ in~\eqref{eq:lap}, defined through
\[
    H_0(f, g, X) = \sum_{i\in[d]} \partial_i f(x) \partial_i g(x),
\]
where $\partial_i$ denote the partial derivative with respect to the $i$-th variables, formally $\partial_i = \partial_{e_i}$ with $e_i$ the $i$-th vector of the canonical basis in $\R^d$.

\paragraph{Reproducing kernel Hilbert spaces.}
In many cases, the spectral functions of an operator $\cL$ are known to belong to a restricted set of functions $\cH$.
For example, spectral functions of most differential operators are expected to belong to the Sobolev spaces $H^{s}$ for many $s$ \citep{Shubin1987}.
Moreover, the set $\cH$ can often be endowed with a Hilbertian structure that makes evaluation map $k_x^*:f\mapsto f(x)$ bounded.
In this case, $k_x^*$ can be represented with $k_x\in\cH$ as $k_x^*(f) = \scap{f}{k_x}_{\cH}$ which is the basis of reproducing kernel \citep{Scholkopf2001,berlinet2011reproducing}.
Typical examples are provided by polynomials of degree less than~$s$, by the class of Sobolev functions $H^{(d+1)/2}$, or by some subspace of analytical functions, for which one can consider respectively
\begin{equation}
    \label{eq:dot-product}
    k_x(y) = q(x^\top y),\quad \text{with}\quad q(t) = (1+t)^s,
\end{equation}
and with $\beta=1$ and $\beta=2$ respectively,
\begin{equation}
    \label{eq:invariant}
    k_x(y) = q(\norm{x-y}),\quad \text{with}\quad q(t) = \exp(-t^\beta).
\end{equation}
The space $\cH$ equates the closure with respect to its inner product of the span of the $k_x$ for $x\in\X$.

\paragraph{Low-dimensional approximation.}
Reproducing kernel Hilbert spaces are usually infinite-dimensional.
However, when dealing with a finite number of data, they can be restricted on a finite-dimensional space as a consequence of representer theorems \citep{Scholkopf2001representer}.
More generally, most of the action on $\cH$ can be restricted to the span of a few number of elements $k_{x_i}$ for $x_i \sim \rho$ \citep{Williams2000}, or to the span of few ``random features'' \citep{Rahimi2007}, which leads to strong computational gain for a low statistical cost \citep{Rudi2015,rudi2021generalization}.
Following the first idea, we set
\begin{equation}
    \label{eq:Nystrom}
	\phi_i = \psi_i = k_{x_i},
\end{equation}
for $(x_i)_{i\in[p]}$ subsampled from the data $(x_i)_{i\in[n]}$, i.e., $p < n$.
The difference between the projections of spectral functions on $\ima F = \Span\{k_{x_i}\}$ and their projections on $\cH$ is naturally characterized from the difference between the covariance $\Sigma =\E[k_Xk_X^*]$ and its empirical estimate $\hat\Sigma$ from the points $(x_i)_{i\in[p]}$, which, when the former has a rapidly decaying spectrum, can vanish quite fast as $p$ increases.

\paragraph{Implementation.}
When considering the objective~\eqref{eq:differential} with the space defined in~\eqref{eq:Nystrom}, $\hat L$ becomes
\[
    \hat L_{ij} = \sum_{k,\alpha,\beta} c_{\alpha,\beta} \partial_\alpha k_{x_i}(x_k) \partial_\beta k_{x_j}(x_k). 
\]
Often, the structure of this matrix allows to reduce, at the cost of an increase in memory space, the number of flops needed to build it compared to a naive implementation.
The next section illustrates this fact through Proposition~\ref{thm:imp} for the specific case where $\cL = \cL_0$.

\section{THE LAPLACIAN EXAMPLE}
\label{sec:lap}
This section details previous considerations for the specific example where $\cL = \cL_0$ in~\eqref{eq:lap}.

\paragraph{Specificities.}
First of all, $\cL_0$ is self-adjoint positive, inheriting its symmetry and positiveness from~$H_0$ in~\eqref{eq:lap}.
Moreover, under mild assumptions on~$\rho$, the Rellich–Kondrachov embedding theorem holds true, which implies the compactness of~$\cL^{-1}$ \citep{Kondrachov1945certain}.
This is useful to apply the spectral theorem and consider the countable eigenvalue decomposition of~$\cL_0$.
Finally, because $\cL_0$ is a diffusion operator, in some cases of interest, its eigenfunctions are polynomials \citep{Bakry2021}, e.g., when $\rho = \cN(0, I)$ or when $\rho$ is uniform on the sphere $\bS^{d-1}$.

\paragraph{Statistical efficiency.}
We now turn to the statistical efficiency of our estimator.
The next theorem, proven in Appendix~\ref{app:stat-L0}, shows how our algorithm succeeds in leveraging the smoothness of spectral functions to guarantee fast rates of convergence.

\begin{theorem}
    \label{thm:stat-L0}
    Assume that $\|x\| \leq M$ almost surely, i.e., $\rho$ has compact support, and that $\cL_0$ is diagonalized by polynomials of increasing order. 
    For $s\in\N$, consider the kernel $k_x(y) = (.5+x^\top y / 2M^2)^s$ and the search for eigenfunctions with~\eqref{eq:Nystrom}.
	For any $\delta\in(0,1)$, there exists a constant $c_\delta$ such that for $n$ large enough, $p = n^{1/2}$ and $s = s_n$ (set so $d\log(\lambda_s)+\log(s) = \log(n)/2$), with probability $1-\delta$, Algorithm~\ref{alg:galerkin} with $\phi_i = \psi_i = k_{x_i}$ guarantees
    \[
	  \big\|\cL^{-1} - G\hat L^{-1} F^*\big\| \leq c_\delta n^{- d/2(d+1)}.
    \]
\end{theorem}

This theorem should be compared with the study of \citet{PillaudVivien2023} that gives rates in $O(n^{-1/4})$ with an algorithm based on a representer theorem that leads to an implementation in $O(n^3d^3)$ flops and $O(n^2d^2)$ memory bits.
In contrast, we can ensure better rates for only $O(n^2 + n^{3/2}d)$ flops and $O(n^{3/2})$ bits.\footnote{%
    To see this, plug $p = n^{1/2}$ into the complexity bounds in $O(np^2+npd)$ and $O(np)$ of the next paragraph.
}
It should equally be compared with graph-Laplacian approaches whose error scales in $O(n^{-1/d})$ for at least $O(n^2p + n^3)$ flops and $O(n^2)$ bits to build and get the eigen decomposition of the graph-Laplacian matrix (which we actually reduce to $O(n^2p + np^2 + npd)$ flops and $O(n^2 + np)$ bits in Appendix~\ref{app:imp}).
Although the result of Theorem~\ref{thm:stat-L0} might seem to break the curse of dimensionality, it should be noted that it could hide constant that grows exponentially fast with respect to the dimension, which may limit the application of our method to large dimensional problems where it is impossible to get a good estimation of $\rho$ without a large number of data \citep{cabannes2023samples}.

\paragraph{Implementation.}
We conclude this section by showing that our method can be implemented with $O(np^2 + npd)$ flops and $O(p^2 + nd)$ memory bits.

\begin{proposition}
  \label{thm:imp}
  Assume that $\X$ is endowed with a scalar product.
  Given a kernel $k_x(y) = q(\norm{x - y})$ defined from $q:\R\to\R$, for $x, y, z\in \X$, we have
  \begin{align}
     &H_0(k_y, k_z, x) = \scap{\nabla k_y(x)}{\nabla k_z(x)}
     \nonumber
     \\&\qquad= \frac{q'(\norm{x-y})}{\norm{x-y}}\frac{q'(\norm{x-z})}{\norm{x-z}}\, (x-y)^\top (x-z).
  \end{align}
  Similarly for dot-product kernel $k_x(y) = q(x^\top y)$, 
  \begin{equation}
	H_0(k_y, k_z, x) = q'(x^\top y)q'(x^\top z) \, y^\top z.
  \end{equation}
\end{proposition}
\begin{proof}
  The proof follows from the application of the chain rule in the calculation of $\scap{\nabla_x k_y(x)}{\nabla_x k_{z}(x)}$.
\end{proof}

Our implementation computes: {\em (i)} $X = (x_i^\top x_k)_{ik} \in \R^{p\times n}$ with $O(npd)$ flops and $O(np)$ bits; {\em (ii)} $q(X)$ and $q'(X)$, where the operator is understood elements wise, with $O(np)$ flops and $O(np)$ bits; {\em (iii)} $\Psi$ and $L$ from $q(X)$, $q'(X)$ and $X$ with $O(np^2)$ flops and $O(np)$ bits; {\em (iv)} the generalized eigen decomposition (GEVD) associated with $(L, \Psi)$ with $O(p^3)$ flops in $O(p^2)$ bits.\footnote{%
    In practice, one might regularize the system as $(L+\epsilon I, \Psi)$ or $(L, \Psi+\epsilon I)$ for a small regularizer $\epsilon > 0$ to avoid diverging solution $\alpha_i = +\infty$ due to inversion instability, especially if the eigenfunctions of $\cL$ do not belong to our parametric models.
}
Adding the steps leads to a total of $O(npd + np^2 + p^3)$ flops and $O(np)$ bits.

\begin{algorithm}[ht]
    \KwData{Data $(x_i) \in \X^n$, kernel $k_x(y) = q(x^\top y)$.}
    Compute $X = (x_i^\top x_j) \in \R^{p\times n}$ with $p \leq n$;\\ 
    Compute $\Psi = q(X) q(X)^\top \in \R^{p\times p}$ elementwise;\\
    Compute $L = (q'(X) q'(X)^\top)$;\\
    Update $L_{ij} \leftarrow X_{ij} L_{ij}$ for all $i, j\in[p]$;\\
    Solve $(\hat\lambda_i, (\alpha_{ij})_{j\in[p]})_{i\in[p]} \leftarrow \operatorname{GEVD}(L, \Psi)$;\\
    Set $\hat f_i(x) := \sum_{j\in[p]} \alpha_{ij} k_{x_j}(x)$.\\
    \KwResult{Estimate $(\hat \lambda_i, \hat f_i)$ of the decomposition of $\cL_0$.}
    \caption{$\cL_0$ estimate with dot-product kernel}
    \label{alg:klap-dot}
\end{algorithm}
\vspace{-1.5em}
\begin{algorithm}[ht]
    \KwData{Data $(x_i) \in \X^n$, kernel $k_x(y) = q(\norm{x-y})$.}
    Compute $X = (x_i^\top x_j), \in \R^{p\times n}$, $D = (x_i^\top x_i) \in \R^n$;\\
    Deduce $N = (\norm{x_i - x_j}) \in \R^{p\times n}$ and $T = q'(N) / N$;\\
    Initialize $L = 0 \in \R^{p\times p}$;
    $\Psi$ = $q(N) q(N)^\top \in \R^{p\times p}$;\\
    \For{$k \in [n]$}{
    Set $\gamma^{(k)}_{ij} := (D_k - X_{ik} - X_{jk} + X_{ij})$;\\
    Update $L_{ij} \leftarrow L_{ij} + \gamma_{ij}^{(k)} T_{ik} T_{jk}$;\\
    }
    Solve $(\lambda_i, (\alpha_{ij})_{j\in[p]})_{i\in[p]} \leftarrow \operatorname{GEVD}(L, \Psi)$;\\
    Set $f_i(x) := \sum_{j\in[p]} \alpha_{ij} k_{x_j}(x)$.\\
    \KwResult{Estimate $(\lambda_i, f_i)$ of the decomposition of $\cL$.}
    \caption{$\cL_0$ estimate with distance kernel}
    \label{alg:klap}
\end{algorithm}
\vspace{-1em}

\section{RELATED APPROACHES}

\subsection{Graph Laplacians}
Graph Laplacians are the classical way to estimate spectral decompositions in machine learning \citep{Zhu2003,Belkin2003,zhu2021contrastive,zhu2022generalized}.  
Although there exist many variants, those methods mainly consist in approximating the Laplacian operator ${\cal L}_0$ with finite differences. 
For $f:\X\to\R$
\begin{equation}
    \label{eq:graph-lap}
    \E[\norm{\nabla f(X)}^2] \approx \sum_{i,j\in [n]} w_{ij} (f(x_i) - f(x_j))^2,
\end{equation}
where the $w_{ij}$ are a set of weights usually taken as $(w_{ij}) = D^{-1/2}\tilde{W}D^{-1/2}$ where $\tilde{w}_{ij} = \exp(-\alpha\|x_i - x_j\|^2)$ with $\alpha$ a scale parameter, and $D$ is the diagonal matrix with $D_{ii} =\sum_{j\in[n]}\tilde w_{ij}$.

Graph Laplacians differ from our approaches in several aspects.
On the positive side, they could present computational advantages, when the evaluation of $H_0$ in~\eqref{eq:lap} is too costly.
Moreover, by tuning the weighting scheme $w$, graph Laplacians can estimate different Laplacians, such as the Laplace-Beltrami operator associated with the data manifold even if the data are non uniform on this manifold \citep{Coifman2006,Hein2007}, while this operator usually can not be written under the form needed to apply our framework from~\eqref{eq:energy}.
However, approximating a differential operator with finite differences is known to be statistically inefficient in high dimension \citep{Bengio2006,Singer2006,Hein2007}, as it will not easily adapt to the smoothness of the targeted operator.
Finally, graph-Laplacians are used to approximate Laplacian operators, while our method is more generic.

\begin{figure*}[t]
    \vspace{-.5em}
    \centering
	\includegraphics{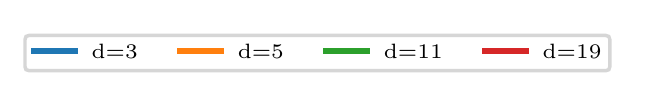}\\
    \vspace{-.5em}
	\includegraphics{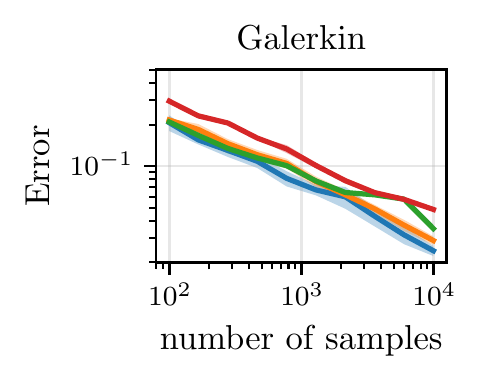}
	\includegraphics{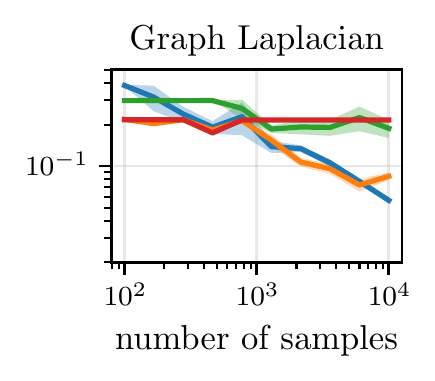}
        \hspace{2em}
	\includegraphics{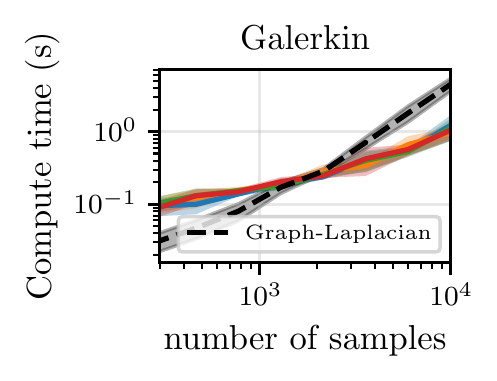}
    \vspace{-.5em}
	\caption{
	  (Left) Testing error~\eqref{eq:sur} when learning the first 25 ``spherical harmonics'' eigenvalues as a function of the number of samples $n$ in different dimension $d$ with Galerkin method.
        (Middle) Same figure with graph-Laplacian.
        The error is averaged over 100 runs, with standard deviations shown in solid color, and we pick the best result over three kernels with five different parameters each with five different values for $p$ (best of 75 for Galerkin), as well as six different scales for weighting in graph-Laplacian (best of 450 for graph-Laplacian).
        (Right) Computation time for Galerkin method with polynomial kernel of degree three and $p=177$.
	  Experimental setups and reproducibility specifications are detailed in Appendix~\ref{app:experiments}.
	}
    \label{fig:eigen}
\end{figure*}

\subsection{Methods based on Loss Optimization}
\label{sec:nn}

In the era of deep learning, one of the main challenges of machine learning pipelines is to design a principled loss, before finding an architecture and an optimizer which can practically minimize this loss when choosing a good set of hyperparameters.
This section explores the core principles beyond our approach, and details the possibility to learn spectral decompositions with any models, including deep neural networks.
To ease the discussion, we assume that $\cL$ is symmetric and thus that $f_i = g_i$ in the following.

In this paper, we started from an objective term $\cE_{\cL}:L^2(\rho)\to\R$, whose minimization of $\sum_{i\in[p]} \cE_{\cL}(f_i)$ under the constraints that the $(f_i)$ are orthogonal in $L^2(\rho)$ retrieves the first eigenspace of $\cL$, viz., the span of the $f_i$ equates the one of the first eigenfunctions $(f_i^*)_{i\in[p]}$ of $\cL$~\eqref{eq:spe-dec}.
This property holds true from any unitary norm \citep{Mirsky1960}, leading to many losses that could be used to train a neural network.

Second, we needed a way to enforce orthogonality between the different $f_i$'s, which led us to consider generalized singular value decomposition.
When the $f_i$'s are learned with optimization methods, one might naively add a regularizer to the objective that reads, with $f:\X\to\R^p$ whose coordinates are the~$f_i$'s,
\begin{align}
    \nonumber
    \cR(f) &= \norm{\E[f(X)f(X)^\top] - I}^2 
    \\&= \E[(f(X)^\top f(X'))^2] - 2\E[\norm{f(X)}^2] + p,
    \label{eq:ortho}
\end{align}
where $X'$ is another independent realization of $X$, the last equation being useful in order to get unbiased estimates from small batches of this objective.
Minimizing the resulting loss $\cE_\cL + \cR$ allows to retrieve the spectral decomposition of $\cL$, since
\begin{align}
    &\argmin_{f:\X\to\R^p} 2\sum_{j\in[p]}\mu_i^{-1}\cE_\cL(f_j) + \cR(f) 
    \nonumber
    \\[-.3cm]&\qquad\qquad\qquad\qquad = \bigg(\sqrt{\Big(1 - \frac{\lambda_i}{\mu_i}\Big)_+} f_i\bigg)_{i\in [p]}.
    \label{eq:ssl-obj}
\end{align}
A proof of this statement can be found in \citet{Zhang2022} \citep[see also][]{johnson2023contrastive,Cabannes2023}.

\subsection{Modeling of Self-Supervised Learning}

Many representation learning methods can be modeled as estimating the first spectral functions of some operators which are compatible with our framework.
In the pre-deep learning area, the operator $\cL_0$ in~\eqref{eq:lap} was the operator of choice to extract ``spectral embeddings''.
Several self-supervised learning algorithms can be seen as using variations of this operator.
First, \citet{Simard1991} suggested learning features that are invariant to small perturbations locally by working with what we would call today ``Jacobian vector products''.
It fell within our framework with
\begin{equation}
  \label{eq:tangent}
  \cE_{\cL_{\rm TP}}(f) = \E_X \E_{U}\bracket{\scap{\nabla f(x)}{U}^2\midvert X},
\end{equation}
where the distribution $(U\,\vert\, X=x)$ specifies the direction of invariance to be enforced at the point $x$.
When $U$ is uniform on the sphere, we retrieve $\cL_0$, although this estimator will suffer from high variance as detailed in Appendix~\ref{app:var}.

More recently, invariance has been enforced by looking at finite differences between different data augmentations $\xi, \xi' \in \R^d$ of the same original image $x$ \citep[e.g.,][]{Chen2020}.
With our formalism, this would be written 
\begin{equation}
  \label{eq:vicreg}
  \cE_{\cL_{\rm SSL}}(f) = \E_X \bracket{\E_{\xi,\xi'}\bracket{\norm{f(\xi) - f(\xi')}^2 \midvert X}},
\end{equation}
if using the square loss to enforce equality between the representations $f(\xi)$ and $f(\xi')$ of the two versions $\xi$ and $\xi'$ obtained from $X$ after data augmentation.\footnote{%
    In the literature, $\cL$ is often rewritten as the integral operator associated with the kernel $p(\xi, \xi')/\sqrt{p(\xi)p(\xi')}$ assuming that $p(\xi) = p(x)$, with $p$ denoting the different densities against the Lebesgue measure (assumed to exist) \citep{Haochen2021,Lee2021,deng2022neural}.
}
In addition to the energy term $\sum_i \cE_{\cL}(f_i)$, self-supervised losses enforce the features $(f_i)$ to differ from one another, either through the use of contrastive pairs \citep{Chen2020} that can be understood through the second equation of~\eqref{eq:ortho} \citep{Haochen2021}, or through the use of a ``whitening regularizer'' \citep{Ermolov2021,Zbontar2021,Bardes2022} that assimilates to the first equation of~\eqref{eq:ortho} \citep{Balestriero2022}.
In other terms, those approaches could be modeled through the loss~\eqref{eq:ssl-obj} and understood as learning spectral embeddings.
Similarly, vision-language models with contrastive language-image pre-training (CLIP) could be modeled with an asymmetric operator defined as per~\eqref{eq:energy}.

\section{EXPERIMENTS}

\subsection{Spherical Harmonics}
In order to check the validity of our methods, we experiment in settings where the ground truth is known.
To this end, let us consider the sphere $\cX = \cS^{d-1}$ in $\R^d$ with the uniform distribution.
The operator $\cL_0$ in~\eqref{eq:lap} identifies to the square of the orbital angular momentum \citep{Condon1935,frye2012spherical}, whose eigenfunctions are known to be the spherical harmonics.
Those are polynomials of increasing order.
In particular, there are $N(d, s)$ independent polynomials of degree $s$, each of them associated with the eigenvalues $\lambda_p = \mu_s$ where, as proven by \citet[][Theorem 4.4 and Proposition~4.5]{frye2012spherical}
\[
    \mu_s = s (s+d-2), \, N(d,s)=\frac{2s+d-2}{s} \binom{s+d-3}{s-1}.
\]
Figure~\ref{fig:spherical} illustrates how our Galerkin approach enables the learning of spherical harmonics.
In order to evaluate the quality of our method, because the operator norm in $L^2(\rho)$ cannot be computed empirically, we use the surrogate metric
\begin{equation}
    \label{eq:sur}
    \cE_S(\hat \lambda) \propto \sum_{i\in[k]} \vert\lambda_i^{-1} - \hat\lambda_i^{-1}\vert, \qquad \cE_S(0) = 1,
\end{equation}
for $k=25$, which is bounded by $k\|\cL^{-1} - G^*L^{-1}F\|$ when the retrieved eigenfunctions $(\hat f_i)$ are orthogonal in $L^2(\rho)$ as a consequence of Weyl's theorem \citep{Weyl1912}.

The left of Figure~\ref{fig:eigen} shows how the eigenvalues retrieved by our method lead to an error $\cE_S \approx c n^{1/2}$, since we go from $\cE_S \approx 2$ to $\cE_S \approx 0.2$ when going from $n=10^3$ to $n=10^5$, independently of the dimension (although the constant $c$ increases as illustrated by the offset of the red curve).
In contrast, as showcased by the middle of Figure~\ref{fig:eigen}, graph Laplacians suffer from the curse of dimensionality.
The right of Figure~\ref{fig:eigen} equally shows how the computation time scales more or less linearly with $n$ when $p$ is given and does not depend much on  dimension.\footnote{%
    The left part of the plot actually shows a better scaling then linear which can be explained by the fact that we are plotting $t_n \simeq c_1 np^2 + c_2 pnd + c_3 p^3$ with $c_3$, due matrix inversion, expected to be much bigger than $c_1$ and $c_2$, due to matrix multiplication.
}
A more thorough discussion on our experimental setup, on our removal of confounders, and on the effect of the different parameters at play is provided in Appendix~\ref{app:experiments}.

\subsection{Hermite Regression}
This paper presents a novel method which might find several applications \citep[e.g.,][for semi-supervised learning]{Cabannes2021}.
For example, it opens the path for Hermite interpolation \citep{Hermite1877}, where given a set of $n$ data points $(x_i)$, one tries to learn a function $f$ that interpolates both $f(x_i) = y_i$ and $\nabla f(x_i) = t_i$ for some known scalar values $(y_i)$ and vectors $(t_i)$ in $\R^d$.
Hermite interpolation is usually approached with the least squares problem,
\[
    \argmin_{f\in\cF} \sum_{i\in[n]}\Big\{ (f(x_i) - y_i)^2 + \norm{\nabla f(x_i) - t_i}^2 \Big\},
\]
with $\cF$ some search space of functions from $\R^d$ to $\R$.
Depending on $\cF$, the argument of the minimum might not interpolate the data and their derivatives, in which case the method could be called Hermite regression.
Notice that the loss we are considering can be rewritten as
\[
    \norm{f(x) - y}^2 + \norm{\nabla f(x) - t}^2 = \norm{Df - z}^2,
\]
where, with $(t_i)$ now denoting the coordinates of $t\in\R^d$,
\[
    Df = (f, \partial_1 f, \dots \partial_d f)^\top, \quad z = (y, t_1, \dots, t_d) \in \R^{d+1}.
\]
In other terms, Hermite regression is a instance of the more generic type of linear regression problems, i.e.,
\[
     \argmin_{f:\X\to\R} \E_{(X,Z)}[\norm{Df(X) - Z}^2] = (D^*D)^{-1} D^* f_z,
\]
with $f_z(x) = \E[Z\,|\,X=x] \in \R^d$ and the adjunction understood in $L^2(\R^d,\R^d,\rho)$ with the $\ell^2$-product topology.

Classical approaches based on representer theorems typically require $O(n^3d^3)$ flops \citep{Zhou2008}.
In contrast, for Hermite regression, we have already seen how to build an approximation of the operator $D^*D = \cL_0 + I$ with $O(np^2 + npd)$ flops.
An approximation of the second term $D^*f_z$ is easier to build, and a naive implementation leads to $O(npd)$ flops.
The resulting algorithms, Algorithm~\ref{alg:hermite-dot} and~\ref{alg:hermite} in Appendix~\ref{app:hermite}, cut down cost to $O(np^2 + npd)$ flops, matching kernel ridge regression implementations that can ``handle billions of points effectively'' \citep{Meanti2020}.
In other terms, we cut down the computational bottleneck associated with the usage of derivatives in kernel methods.

\begin{figure}
    \vspace{-.5em}
    \centering
    \includegraphics{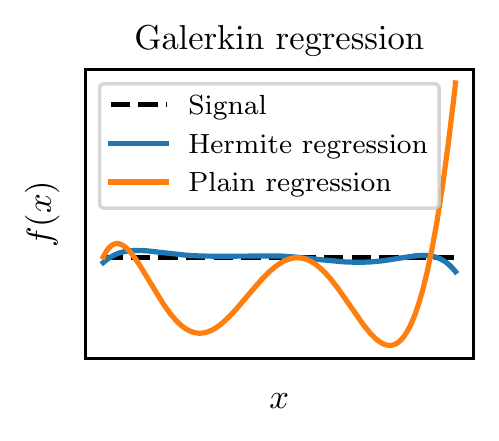}
    \vspace{-.5em}
    \caption{
    Comparison between plain regression and ``Hermite regression'' with the Gaussian kernel, $n=1000$ and $p=100$ when learning a constant function without noise (a task known to be hard for the Gaussian kernel).
    }
    \label{fig:hermite-inter}
    \vspace{-1em}
\end{figure}

\section{CONCLUSION}

This paper introduced an algorithm that can be seen as an instance of the Galerkin method to compute the spectral decomposition of a large class of operators.
It showcased its usefulness theoretically through a series of approximation guarantees.
It was put in perspective with graph-Laplacians, which can be seen as estimating differential operators with finite differences, and suffer from the curse of dimensionality.
Those statistical considerations were validated empirically.
We later detailed efficient implementations of our approach with structured reproducing kernels, which we have packaged into a Python library to be used off-the-shelf.
Those efficient implementations break down computational bottlenecks arising when dealing with derivatives in kernel methods based on representer theorems.
In particular, we show how one may perform Hermite regression with $O(np^2 + npd)$ flops, with $p$ chosen by the user, instead of a naive implementation that would scale in $O(n^3d^3)$.

Finally, we discussed the core principle beyond our approach to design losses whose optimization enables learning the spectral decomposition of linear operators with non-linear spaces of functions, e.g., with deep neural networks.
Those losses were designed to be convex on the cone of positive matrices ($f(X)f(X)^\top$), although when models are not convex, training dynamics might exhibit robustness and stability issues, requiring proper hyperparameters tuning to induce behaviors of interest.
We equally extended on how our setting models recent approaches to representation learning, at least when losses are approximated with squared distances, unveiling abstract linear operators beyond them.

%% file: appendix/proof.tex
\paragraph{Limitations.}
It should be noted that, although it is likely, it has not been formally proven that graph Laplacians suffer from minimax lower bound in $\Omega(n^{1/d})$ (or alike) over the class of operators with smooth eigen-functions.
Moreover, even if such lower bounds were proved, it does not rule out the highly unlikely possibility that there exists a single operator for which some graph Laplacian method convergences in $\Theta(n^{1/d})$ while our method does in $\Theta(n^{1/2})$ (or alike), while for all the other operators it beats our method.
Finally, note that one can always define the class of functions for which a method with tuned hyper parameters "beats the curse of dimensionality". 
However, for graph Laplacians, these classes of functions are likely to be of low interest regarding operators encountered in real problems.

\section{Proofs}

\subsection{Characterization of \texorpdfstring{$\cL_0$}{the Laplacian}}
\label{app:char-L0}

The characterization of $\cL_0$ follows from multidimensional integration by part,
\begin{align*}
    \scap{f}{\cL_0 g}_{L^2(\rho)} 
    &= \int f(x) (\cL_0 g)(x) p(x)\diff x 
    = \E[\scap{\nabla f(X)}{\nabla g(X)}]
    \\&= \lim_{r\to+\infty} \sum_{i\in[d]} \int_{\norm{x}< r} \underbrace{\partial_i f(x)}_{u'} \underbrace{p(x) \partial_i g(x)}_{v}\diff x 
    \\&= \lim_{r\to+\infty} \sum_{i\in[d]} \int_{\norm{x}=r} \underbrace{f(x)}_{u}\underbrace{p(x)\partial_i g(x)}_{v} \scap{e_i}{n} \diff S  - \int_{\norm{x}< r} \underbrace{f(x)}_{u} \underbrace{\partial_i (p(x) \partial_i g(x))}_{v'}\diff x 
    \\&=-\int f(x) \frac{1}{p(x)}\sum_{i\in[d]} \partial_i (p(x) \partial_i g(x)) p(x)\diff x,
\end{align*}
where we used the fact, when $p$ is regular enough, e.g. $V$ is coercive, the surface integral goes to zero when $g$ and $f$ are smooth enough.
As a consequence,
\begin{align*}
    (\cL_0 f)(x)
    &= -\frac{1}{p(x)}\sum_{i\in[d]} \partial_i (p(x) \partial_i f(x))
    = -\frac{1}{p(x)}\sum_{i\in[d]} p(x) \partial_{ii}^2 f(x) + \partial_i p(x) \partial_i f(x)
    \\&= - \Delta f(x) - \frac{1}{p(x)}\scap{\nabla p(x)}{\nabla f(x)}
    = - \Delta f(x) - \scap{\nabla \log p(x)}{\nabla f(x)}
    \\&=  - \Delta f(x) + \scap{\nabla V(x)}{\nabla f(x)}.
\end{align*}

\subsection{Operator Details}
\label{app:operator}
Consider the mapping $F$, we have
\[
    \scap{f}{F\alpha}_{L^2(\rho)} = \sum_{i\in[p]} \alpha_i \scap{f}{\phi_i}_{L^2(\rho)} = \scap{(\scap{f}{\phi_i}_{L^2(\rho)})_{i\in[p]}}{\alpha}_{\R^p} = \scap{F^*f}{\alpha}_{\R^p}.
\]
As a consequence, with the adjunction with respect to $L^2(\rho)$ and $(\R^p, \ell^2)$
\[
    F^*:f\in L^2(\rho) \mapsto (\scap{f}{\phi_i}_{L^2(\rho)})_{i\in[p]} \in \R^p.
\]
We deduce the characterization of $\Phi$ given in \eqref{eq:Phi},
\[
    F^*F\alpha = (\scap{F\alpha}{\phi_i}_{L^2(\rho)})_{i\in[p]} =  (\sum_{j\in[n]}\alpha_j\scap{\phi_j}{\phi_i}_{L^2(\rho)})_{i\in[p]} = (\scap{\phi_j}{\phi_i}_{L^2(\rho)})_{i,j\in[p]}\alpha = \Psi\alpha.
\]
Similarly for the characterization of $L$ in \eqref{eq:L},
\[
    F^*\cL G\alpha = (\scap{\cL G\alpha}{\phi_i}_{L^2(\rho)})_{i\in[p]}= (\cE_\cL(G\alpha, \phi_i))_{i\in[p]} =  (\sum_{j\in[n]}\alpha_j\cE_\cL(\phi_j,\psi_i))_{i\in[p]} = (\cE_\cL(\phi_j,\psi_i))_{i,j\in[p]}\alpha = L\alpha.
\]

\subsection{Generalized Singular Value Decomposition}
\label{app:gsvd}

Let us consider three matrices $L, \Phi, \Psi$ in $\R^{p\times p}$.
We would like to show that there exists three matrices $A, B, S$ such that
\[
    A L B^\top = S, \qquad A\Phi A^\top = I, \quad B\Psi B^\top = I.
\]
Remark that this generalized SVD is not the two-matrices version of \citet{vanLoan1976}, but the weighted single-matrix one that appeared in correspondence analysis \citep{Greenacre1984,Jolliffe2002}.

To find such a decomposition, let us consider the singular value decomposition of 
\[
    \Phi^{-1/2} L \Psi^{-1/2} = XSY^\top, \qquad X^\top X = Y^\top Y = I.
\]
We have
\[
    X^\top\Phi^{-1/2} L\Psi^{-1/2}Y = S, \qquad (\Phi^{-1/2}X)^\top \Phi (\Phi^{-1/2}X) = (\Psi^{-1/2}Y)^\top \Psi (\Psi^{-1/2}Y) = I.
\]
As a consequence, setting $A = (\Phi^{-1/2}X)^\top$ and $B = (\Psi^{-1/2}Y)$ we get
\[
    A L B^\top = S, \qquad A\Phi A^\top = B\Psi B^\top = I.
\]
The first equation can equally be written, using that $A^{-1} = \Phi A^\top$,
\[
    A L B^\top = S
    \qquad\Leftrightarrow\qquad L B^\top = \Phi A^\top S
    \qquad\Leftrightarrow\qquad L = \Phi A^\top S B \Psi 
    \qquad\Leftrightarrow\qquad A L = S B \Psi.
\]
It can equally be expressed in term of columns as
\[
    [LB^\top]_i = L [B^\top]_i = L\beta_i = [\Phi A^\top S]_i = S_{ii} [\Phi A^\top]_i = S_{ii} \Phi [A^\top]_i = S_{ii} \Phi \alpha_i,
\]
which matches the docstring formulation of Scipy for the symmetric case (with $\alpha_i = \beta_i = {\tt v}$, $S_{ii} = {\tt w}$, $L = {\tt a}$, $\Phi = {\tt b}$):
\begin{center}
    {\small \url{https://github.com/scipy/scipy/blob/v1.11.3/scipy/linalg/_decomp.py#L325}}.
\end{center}
The advantage of using the GSVD rather than the SVD of the system $\Psi^{-1/2}L\Phi^{-1/2}$ is that it requires less flops (although the big-O complexity will be the same).
The GSVD is roughly equivalent to one matrix inversion instead of three if we choose to first invert $\Psi$ and $\Phi$ before performing one SVD \citep{Golub1970}.

\subsection{Proof of Theorem \ref{thm:stat-phi}}
\label{app:stat-phi}

\begin{lemma}[Error decomposition]
  With the $(\hat \lambda_i, \hat f_i, \hat g_i)$ obtained with Algorithm \ref{alg:galerkin}, when $L^{-1}(\hat L - L) - I/2$ is positive, for any $a \in [0,1]$, 
    \[
        \big\| \cL^{-1} - (\sum_{i\in[p]}\hat\lambda_i \hat f_i \otimes \hat g_i)^{-1} \big\|
	  \leq \big\| (I-\Pi_G)\cL^{-1} \| + \| \cL^{-1}(I-\Pi_F) \|
	+ \gamma\norm{\Psi^{1/2}L^{-a}}\norm{L^{-a}\Phi^{1/2}}\norm{L^{-(1-a)}(\hat L - L)L^{-(1-a)}},
    \]
  where $\gamma = 1 + 2(\norm{L}\norm{L^{-1}})^{\min(a, 1-a)}$.
\end{lemma}
\begin{proof}
	We start with
	\[
	  \big\| \cL^{-1} - (\sum_{i\in\N}\hat\lambda_i \hat f_i \otimes \hat g_i)^{-1} \big\|
	  \leq 
	  \big\| \cL^{-1} - (\Pi_F \cL \Pi_G)^{-1} \| 
	  + \big\| (\Pi_F \cL \Pi_G)^{-1} - \sum_{i\in\N} \hat\lambda_i^{-1} \hat g_i \otimes \hat f_i \big\|.
  \]
  The first term is rewritten with
  \begin{align*}
	  \big\| \cL^{-1} - (\Pi_F \cL \Pi_G)^{-1} \| 
	  &= \big\| \cL^{-1} - \Pi_G \cL^{-1} \Pi_F \| 
	  = \big\| (I-\Pi_G)\cL^{-1} + \Pi_G \cL^{-1}(I-\Pi_F) \| 
	  \\&\leq \big\| (I-\Pi_G)\cL^{-1} \| + \| \Pi_G \cL^{-1}(I-\Pi_F) \| 
	  \\&\leq \big\| (I-\Pi_G)\cL^{-1} \| + \| \cL^{-1}(I-\Pi_F) \|.
  \end{align*}
  The second term is rewritten with
  \begin{align*}
	  \big\| (\Pi_F \cL \Pi_G)^{-1} - \sum_{i\in\N} \hat\lambda_i^{-1} \hat g_i \otimes \hat f_i \big\|
	  &= \big\| G(F^* \cL G)^{-1}F^* - \sum_{i\in\N} \hat\lambda_i^{-1} \hat g_i \otimes \hat f_i \big\|
	  = \big\| G(L^{-1} - \hat L^{-1})F^*\big\|
	\\&= \big\| \Psi^{1/2}(L^{-1} - \hat L^{-1})\Phi^{1/2}\big\|.
  \end{align*}
  The difference between the inverse can be worked out as,
  \begin{align*}
	  L^{-1} - \hat L^{-1}
	  &= L^{-1}(\hat L - L)\hat L^{-1}
	  = L^{-1}(\hat L - L)(L^{-1} + \hat L^{-1} - L^{-1})
	  \\&= L^{-1}(\hat L - L)L^{-1} - L^{-1}(\hat L - L) (L^{-1} - \hat L^{-1})
	  \\&= (I + L^{-1}(\hat L - L))^{-1} L^{-1}(\hat L - L)L^{-1},
  \end{align*}
  where the last equality is true when the matrix $I + L^{-1}(\hat L - L)$ is invertible, which is notably implied by $L^{-1}(\hat L - L) \succeq -I$.
  As a consequence,
  \begin{align*}
	\Psi^{1/2}(L^{-1} - \hat L^{-1})\Phi^{1/2}
	&= \Psi^{1/2}L^{-1}(\hat L - L)L^{-1}\Phi^{1/2} + \Psi^{1/2}L^{-1}(\hat L - L)(\hat L^{-1} - L^{-1})\Phi^{1/2}
  \\&=  \Psi^{1/2}L^{-1}(\hat L - L)L^{-1}\Phi^{1/2} + \Psi^{1/2}L^{-1}(\hat L - L)(I + L^{-1}(\hat L - L))^{-1} L^{-1}(\hat L - L)L^{-1}\Phi^{1/2}.
  \end{align*}
  We can translate this last equality in operator norm: for any $a, b\in[0,1]$
  \begin{align*}
	&\norm{\Psi^{1/2}(L^{-1} - \hat L^{-1})\Phi^{1/2}}
	\leq \norm{\Psi^{1/2}L^{-a}}\norm{L^{-b}\Phi^{1/2}}
  \\&\qquad\qquad \times \paren{\norm{L^{-(1-a)}(\hat L - L)L^{-(1-b)}} + \norm{L^{-(1-a)}(\hat L - L)(I + L^{-1}(\hat L - L))^{-1} L^{-1}(\hat L - L)L^{-(1-b)}}}.
  \end{align*}
  The last term can be bounded either with 
  \begin{align*}
  &	\norm{L^{-(1-a)}(\hat L - L)(I + L^{-1}(\hat L - L))^{-1} L^{-1}(\hat L - L)L^{-(1-b)}}
 \\ &	= \norm{L^a L^{-1}(\hat L - L)(I + L^{-1}(\hat L - L))^{-1} L^{-a} L^{-(-1-a)}(\hat L - L)L^{-(1-b)}}
 \\ &\leq \norm{L^a}\norm{L^{-a}} \norm{L^{-1}(\hat L - L)(I + L^{-1}(\hat L - L))^{-1}}\norm{L^{-(1-a)}(\hat L - L)L^{-(1-b)}},
  \end{align*}
  or with,
  \begin{align*}
  &	\norm{L^{-(1-a)}(\hat L - L)(I + L^{-1}(\hat L - L))^{-1} L^{-1}(\hat L - L)L^{-(1-b)}}
 \\ &	= \norm{L^{-(1-a)}(\hat L - L)L^{-(1-b)} L^{1-b}(I + L^{-1}(\hat L - L))^{-1} L^{-1}(\hat L - L)L^{-(1-b)}}
 \\ &\leq \norm{L^{(1-b)}}\norm{L^{-(1-b)}} \norm{L^{-1}(\hat L - L)(I + L^{-1}(\hat L - L))^{-1}}\norm{L^{-(1-a)}(\hat L - L)L^{-(1-b)}}.
  \end{align*}
  Using the fact that $\norm{(I+ A)^{-1} A} \leq 2$ as soon as $A + I/2 \succeq 0$, we deduce that as soon as $L^{-1}(\hat L - L) \succeq - I/2$, 
  \begin{align*}
	&\norm{\Psi^{1/2}(L^{-1} - \hat L^{-1})\Phi^{1/2}}
	\leq \norm{\Psi^{1/2}L^{-a}}\norm{L^{-b}\Phi^{1/2}}\norm{L^{-(1-a)}(\hat L - L)L^{-(1-b)}} \paren{1 + 2(\norm{L}\norm{L^{-1}})^{\min(a, 1-b)}}.
  \end{align*}
  This explains the decomposition in the lemma.
\end{proof}

We continue by bounding the empirical versus population difference $\norm{L - \hat L}$.
To do so, we will use Bernstein concentration inequality.

\begin{lemma}
    \label{lem:estimation}
    Let $A(x)$ be a $p\times p$ matrix bounded by $M$ and $B_1 = \E[A(X)^\top A(X)]$, $B_2 = \E[A(X)^\top A(X)]$. 
	For $t \geq 0$,
    \[
        \Pbb_{(x_k)}\paren{\bigg\|\E[A(X)] - \frac{1}{n}\sum_{k\in[n]}A(x_k)\bigg\| \geq t} 
        \leq 4\frac{\trace{B_1+B_2}}{\max\norm{B_i}}\exp\paren{\frac{-3nt^2}{6\max(\norm{B_i}) + 2Mt}}.
    \]
    For any $t\in[0, M]$,
    \[
        \Pbb_{(x_k)}\paren{\bigg\|\E[A(X)] - \frac{1}{n}\sum_{k\in[n]}A(x_k)\bigg\| \geq t} 
        \leq 2p\exp\paren{\frac{-3nt^2}{8M^2}}.
    \]
\end{lemma}
\begin{proof}
    This is Theorem 7.3.1 and Theorem 6.1.1 of \citet{tropp2015introduction}.
\end{proof}

\begin{lemma}[Estimation error]
  With $n$ data points, our algorithm guarantees, for $t < p^2 H_\infty^2$,
  \[
	\Pbb(\norm{L - \hat L} > t) \leq 2p\exp(-\frac{3nt^2}{8p^2 H_\infty^2}),
  \]
  where $H_\infty = \sup_{i,j,x} H(\phi_i, \psi_j, x)$.
  In other terms, for any $\delta \in (0, 1)$ and $n > 3\log(2p/\delta)$, with probability $1-\delta$
  \[
	 \| L - \hat L \| \leq \sqrt{\frac{8p^2H_\infty^2}{3n}\log(2p/\delta)}.
  \]
\end{lemma}
\begin{proof}
  We would like to use Bernstein inequality with $L$ and $\hat L$, they are both built from the matrix
  \[
	  A(x)_{ij} = H(\phi_i, \psi_j, x).
  \]
  Without specific structure on $H$, we proceed with the following bound
  \[
	\norm{A(x)}^2_{\rm op} \leq \norm{A(x)}^2_{2} \leq \sum_{i,j} H(\phi_i, \psi_j, x)^2 \leq p^2 \sup H(\phi_i, \psi_j, x)^2 =: p^2 H_\infty^2.
  \]
  As a consequence, we have the following bound
  \[
	\Pbb(\norm{L - \hat L} > t) \leq 2p\exp(-\frac{3nt^2}{8M^2}), \qquad\text{with}\qquad M = p^2H_\infty^2.
  \]
  To parse the bound more easily, let us invert $t$, we set
  \[
	  \delta = 2p\exp(-\frac{3nt^2}{8M^2}),\qquad
	  t = \sqrt{\frac{8M^2}{3n}\log(2p/\delta)}.
  \]
  When $n$ is big enough to ensure $t < M$, i.e., $n > 3 \log(2p/\delta)$, we have that with probability $1-\delta$
  \[
	  \norm{L-\hat L} \leq \sqrt{\frac{8M^2}{3n}\log(2p/\delta)}.
  \]
  Replacing $M$ by $pH_\infty$ ends the proof.
\end{proof}

The proof of the theorem follows directly from the previous lemmas.

\subsection{Proof of Theorem \ref{thm:stat}}
\label{app:stat}

In order to prove Theorem \ref{thm:stat}, we need to specify the values taken by $\norm{\cL(I-\Pi_F)}$ and $\norm{(I-\Pi_G)\cL}$.
In all the following, it is useful to introduce 
\begin{equation}
  \hat\Sigma = p^{-1}\sum_{i\in[p]}\phi_i\phi_i^*, \qquad\text{and}\qquad\hat\Xi = p^{-1}\sum_{i\in[p]}\psi_i \psi_i^*.
\end{equation}

\begin{lemma}
   Let $\cA$ be some operator in $L^2(\rho)$.
   Assume that the $(\phi_i)_{i\in[p]}$ were chosen independently at random according to the same distribution $\mu\in\prob{L^2(\rho)}$, with $\Sigma = \E_{\phi\sim\mu}[\phi\phi^*]$.
   Assume that there exists $a\in[0,1]$ and $M > 0$ such that $\norm{\Sigma^{-a/2}\phi}_{L^2(\rho)} \leq M^{1/2}$ independently of $\phi$.
    Let $\Pi_*$ denotes a $\kappa$-dimensional projection on an eigenspace of $\Sigma$ and $\Sigma_*$ the associated restriction $\Sigma_* = \Pi_*\Sigma$.
   For any $\delta \in (0, 1)$ for any $p > 3\log(2\kappa/\delta)$, it holds with probability $1-\delta$, 
    \[
	  \norm{(I-\Pi_F)\cA} \leq \norm{(I-\Pi_*)\cA} + \norm{\Sigma_*^{-(1-a)}\cA} \sqrt{\frac{8M^2}{3p}\log(2\kappa / \delta)}.
    \]
    Here, $A^{-1}$ denotes the pseudo-inverse of $A$.
\end{lemma}
\begin{proof}
  For simplicity, we denote $\Pi = \Pi_F$ in the proof.
  We split the error with
  \[
	\norm{(I-\Pi)\cA} = \norm{(I-\Pi)(I-\Pi_*)\cA + (I-\Pi)\Pi_*\cA}
	\leq \norm{(I-\Pi_*)\cA} + \norm{(I-\Pi)\Pi_* \cA}.
  \]
  The first term depends on assumptions on the problem, while the second term depends on the empirical approximation of $\Pi_*$ by $\Pi$.
  Using the fact that, with $A^{\dagger}$ denoting the pseudo-inverse,
  \[
	A^\dagger - B^\dagger = A^\dagger(B-A)B^{\dagger} - (I-\Pi_A) B^\dagger + A^\dagger (I-\Pi_B),
  \]
  we have
  \begin{align*}
	\Pi_*-\Pi 
	&= \Sigma_*^{\dagger}\Sigma - \hat\Sigma^{\dagger}\hat \Sigma
	= \Sigma_*^{\dagger}(\Sigma - \hat\Sigma) + (\Sigma_*^{\dagger} - \hat\Sigma^{\dagger})\hat \Sigma
  \\&= \Sigma_*^{\dagger}(\Sigma - \hat\Sigma) - \Sigma_*^{\dagger}(\Sigma - \hat\Sigma) \hat\Sigma^{\dagger}\hat \Sigma
	 - (I-\Pi_*) \hat\Sigma^{\dagger}\hat \Sigma
	 + \Sigma_*^{\dagger}(I-\Pi)\hat \Sigma.
  \\&= \Sigma_*^{\dagger}(\Sigma - \hat\Sigma)(I-\Pi) - (I-\Pi_*) \Pi
  = \Sigma_*^{\dagger}(\Sigma - \hat\Sigma)(I-\Pi) + (\Pi_*-\Pi) \Pi.
  \end{align*}
  As a consequence
  \[
    \cA^*\Pi_* (I-\Pi) =\cA^*(\Pi_* - \Pi)(I-\Pi) = \cA^*\Sigma_*^{\dagger}(\Sigma - \hat\Sigma)(I-\Pi).
    = \cA^*\Sigma_*^{\dagger} \Pi_*(\Sigma - \hat\Sigma)(I-\Pi),
  \]
  which we translate in operator norm with
  \[
	\norm{(I-\Pi)\Pi_*\cA} \leq \norm{\Sigma_*^{-(1-a)}\cA}\norm{\Pi_*\Sigma^{-a}(\Sigma - \hat\Sigma)}.
  \]
  We now need to bound the difference between $\Sigma$ and $\hat\Sigma$.
  This is a simple application of Bernstein with $\Pi_*\Sigma^{-a}\phi(X)\phi(X)^\top$.
  In particular, if $\norm{\Sigma^{-a/2}\phi}_{L^2(\rho)} \leq M^{1/2}$, we have for $t\in[0,M]$,
  \[
	\Pbb(\norm{\Pi_*\Sigma^{-a}(\Sigma - \hat \Sigma)} > t) \leq 2\kappa \exp(-\frac{3pt^2}{8M^2}).
  \]
  Once again, let us invert $t$, we set
  \[
	  \delta = 2\kappa\exp(-\frac{3pt^2}{8M^2}),\qquad
	  t = \sqrt{\frac{8M^2}{3p}\log(2\kappa/\delta)}.
  \]
  When $p$ is big enough to ensure $t < M$, in particular, when $p > 3\log(2\kappa/\delta)$, we have that with probability $1-\delta$
  \[
	  \norm{\Pi_*\Sigma^{-a}(\Sigma - \hat \Sigma)} \leq \sqrt{\frac{8M^2}{3p}\log(2\kappa / \delta)}.
  \]
  This ends the proof of the lemma.
\end{proof}

The proof of the theorem follows directly from the previous lemma.
In the setting of Theorem \ref{thm:stat}, it is interesting to specify the value of $\norm{L^{-a}\Psi^{1/2}}$.

\begin{lemma}
  \label{lem:fake-constant}
  Let $\norm{\psi}_{L^2(\rho)} \leq M^{1/2}$ and $\norm{\Xi^{-1/2}\psi}_{L^2(\rho)} \leq M_\infty$ with $M_\infty > 1/2$.
  For any $a\in[0,1]$ if $L$ is symmetric, or for $a=1$ and any $L$, for any $\delta \in (0, 1)$ and $\kappa\in\N$, when
  \[
	  p \geq \max(11 M_\infty^2\log(2\kappa/\delta), 5M^2\log(\frac{2M\mu_{\kappa+1}}{\norm{\Xi}})),
  \]
  it holds with probability at least $1-\delta$,
  \[
	\norm{L^{-a}\Psi^{1/2}} \leq 2^a M^{(1-a)/2}\paren{\norm{\Xi^{-1/2} \cL^{-1}} + \mu_{\kappa+1}^{-1/2}}^a,
  \]
  where $\mu_{\kappa}$ is the $\kappa$-th eigenvalue of $\Xi$.
\end{lemma}
\begin{proof}
  Note that there exists two isometric mappings $U:L^2(\rho)\to\R^p$ and $V:L^2(\rho)\to\R^p$ such that 
  \[
	F = U\hat\Sigma^{1/2}, \qquad\text{and}\qquad G = V\hat\Xi^{1/2},
  \]
  and $UU^* = VV^* = I$. 
  As a consequence
  \[
	L^{-a}\Psi^{1/2} = (F^*\cL G)^{-a} (F^*F)^{1/2} = V\hat\Xi^{-a/2} \cL^{-a} \hat\Sigma^{-a/2}\hat\Sigma^{1/2}U^*,
  \]
  which we translate in operator norm with
  \[
	\norm{L^{-a}\Psi^{1/2}} = \norm{\hat\Xi^{-a/2} \cL^{-a} \hat\Sigma^{(1-a)/2}} \leq \norm{\hat\Xi^{-1/2}\cL^{-1}}^{a}\norm{\hat\Sigma}^{(1-a)/2},
  \]
  where we have used the fact that $\norm{A^aB^a}\leq \norm{AB}^a$ is $A$ and $B$ are positive \citep{Cordes1987}.
  For the last term, we know that 
  \[
	\norm{\hat\Sigma} \leq \trace(\hat\Sigma) = n^{-1}\sum_{i\in[n]} \norm{\phi_i}^2 \leq M.
  \]
  For the first term, considering $\Pi_*$ a projection on the top $\kappa$ eigen functions of $\Sigma$, we have
  \begin{align*}
	\norm{\hat\Xi^{-1/2}\cL^{-1}}
	&\leq \norm{\hat\Xi^{-1/2}\Pi_*\cL^{-1}}
	+ \norm{\hat\Xi^{-1/2}(I-\Pi_*)\cL^{-1}}
	&\leq \norm{\hat\Xi^{-1/2}\Pi_*\cL^{-1}}
    + \norm{\hat\Xi^{-1/2}(I-\Pi_*)}\norm{\cL^{-1}}.
  \end{align*}
  The first part will concentrate with
  \begin{align*}
	\norm{\hat\Xi^{-1/2}\Pi_*\cL^{-1}}^2
	&= \norm{\cL^{-1}\Pi_*\hat\Xi^{-1}\Pi_*\cL^{-1}} 
	= \norm{\cL^{-1}\Xi^{-1/2}\Xi^{1/2}\Pi_*\hat\Xi^{-1}\Pi_*\Xi^{1/2}\Xi^{-1/2}\cL^{-1}} 
	\\&\leq \norm{\cL^{-1}\Xi^{-1/2}}^2 \norm{\Pi_*\Xi^{1/2}\hat\Xi^{-1}\Xi^{1/2}\Pi_*}. 
  \end{align*}
  We have already seen how to treat the last term
  \begin{align*}
	\Xi^{1/2}\hat\Xi^{-1}\Xi^{1/2} 
	&= I  + \Xi^{1/2}(\hat\Xi^{-1}-\Xi^{-1})\Xi^{1/2} 
	= I + \Xi^{-1/2}(\Xi-\hat\Xi)\Xi^{-1/2}\Xi^{1/2}\hat\Xi^{-1}\Xi^{1/2}
  \\&= (I - \Xi^{-1/2}(\Xi-\hat\Xi)\Xi^{-1/2})^{-1}.
  \end{align*}
  Using Bernstein inequality, we deduce that, if $M_\infty > 1/2$,
  \[
	\Pbb(\norm{\Pi_*\Xi^{-1/2}(\Xi - \hat \Xi)\Xi^{-1/2}} > 1/2) \leq
	2\kappa \exp(-\frac{3p}{32 M_\infty^2}).
  \]
  Let us invert the relation, we want
  \[
     \delta \leq 2\kappa\exp(-\frac{3p}{32M_\infty^2}),\qquad
	  p \geq \frac{32M_\infty^2}{3}\log(2\kappa/\delta).
  \]
  For the second part, we proceed with
  \begin{align*}
	\norm{\hat\Xi^{-1/2}(I-\Pi_*)}^2 
	&= \norm{(I-\Pi_*)\hat \Xi^{-1}(I-\Pi_*)}
	\leq \norm{(I-\Pi_*)\Xi^{-1}(I-\Pi_*)}
    + \norm{(I-\Pi_*)(\hat \Xi^{-1} - \Xi^{-1}(I-\Pi_*)}
  \\&\leq \norm{(I-\Pi_*)\Xi^{-1/2}}^2
	  + \norm{(I-\Pi_*)\Xi^{-1}} \norm{\Xi - \hat \Xi}\norm{\Xi^{-1}(I-\Pi_*)},
  \end{align*}
  which can be rewritten as
  \[
	\norm{\hat\Xi^{-1/2}(I-\Pi_*)}^2
  \leq \norm{(I-\Pi_*)\Xi^{-1/2}}^2\paren{1+ \norm{\Xi - \hat \Xi}\norm{\Xi^{-1/2}(I-\Pi_*)}^2}
  \leq \frac{\mu_{\kappa+1}^{-1}}{1-\mu_{\kappa+1}^{-1}\norm{\Xi - \hat \Xi}}.
  \]
  Using Bernstein inequality in Hilbert space \citep{Minsker2017}, we have
  \[
	\Pbb(\norm{\Xi - \hat \Xi} > t) \leq 2\frac{\trace{\Xi}}{\norm{\Xi}} \exp(-\frac{3pt^2}{8M^2})
	\leq \frac{2M}{\norm{\Xi}} \exp(-\frac{3pt^2}{8M^2}).
  \]
  In particular, when 
  \[
	\frac{2M}{\norm{\Xi}} \exp(-\frac{27p}{128M^2}) \leq \mu_{\kappa+1}^{-1} \qquad \Leftrightarrow\qquad
	\frac{128M^2}{27}\log(\frac{2M\mu_{\kappa+1}}{\norm{\Xi}}) \leq p
  \]
  we have that $\norm{\hat\Xi^{-1/2}(I-\Pi_*)} \leq 2\mu_{\kappa+1}^{-1/2}$.
  Collecting the different pieces proves the lemma.
\end{proof}

\subsection{Proof of Theorem \ref{thm:stat-L0}}
\label{app:stat-L0}

To prove Theorem \ref{thm:stat-L0}, we start by reworking the estimation error between $\hat L$ and $L$.

\begin{lemma}[Estimation error for $\cL_0$]
  When $\cL = \cL_0$ and $\phi_i = \psi_i = k_{x_i}$ with $k_x(y) = ((1+x^\top y) / (1+M))^s$ for $M$ an almost sure upper bound on $\norm{X}$, with $n$ data points, our algorithm guarantees, for $t < s^2$,
  \[
	\Pbb(\norm{L - \hat L} > t) \leq 2\kappa\exp(-\frac{3nt^2}{8s^2}), \qquad\text{with}\qquad\kappa = \binom{d+s}{s}.
  \]
  As a consequence, for any $\delta \in (0, 1)$ and $n > 3 \log(2\kappa/\delta)$, it holds with probability $1-\delta$
  \[
	\frac{\norm{L^{-1}}}{1-\norm{L^{-1}}\|L - \hat L\|}  \| \Phi \|^{-1/2}\| \Psi \|^{-1/2} \| L^{-1} \| \| L - \hat L \| \leq 2 \norm{L^{-1}}^2 \sqrt{\frac{8s^2}{3n}\log(2\kappa/\delta)}.
  \]
\end{lemma}
\begin{proof}
  The proof follows the one of Lemma \ref{lem:estimation} with
  \[
	  A(x)_{ij} = H(\phi_i, \psi_j, x).
  \]
  Without specific structure on $H$, we proceeded with $\norm{A(x)} \leq p H_\infty$.
  In the specific case of $\cL_0$, with $\cF = \cG = \cH$, we get
  \begin{align*}
	\norm{A(x)} 
	&= \sup_{\norm{c}=1} c^\top A(x) c = \sup_{\norm{c}=1} \sum_{ij} c_i \scap{\nabla k_{x_i}(x)}{\nabla k_{x_j}(x)} c_j
  \\&= \sup_{\norm{c}=1}\|\sum_i c_i \nabla k_{x_i}(x)\|^2
  \leq \sup_{\norm{c}=1}\sum_i c_i^2 \norm{\nabla k_{x_i}(x)}^2
  \leq \sup \norm{\nabla k_{x}(y)}^2.
  \end{align*}
  When $\cH$ is defined from the kernel
  \[
	  k_x(y) = \paren{\frac{1+x^\top y}{1 + M}}^s \leq 1,
  \]
  this becomes
  \[
	\norm{A(x)} \leq \sup \norm{\nabla k_x(y)}^2 = \sup s^2\paren{\frac{1+x^\top y}{1+M}}^{2(s-1)} \norm{x/(1+M)}^2\leq s^2.
  \]
  Similarly, this choice of $k$ guarantees
  \[
	\norm{\Phi^{1/2}}^2 = \norm{\Phi} \leq \sup \norm{k_x(y)} \leq 1 .
  \]
  As a consequence, we have the following bound
  \[
	\Pbb(\norm{L - \hat L} > t) \leq 2\kappa\exp(-\frac{3nt^2}{8s^4}),
  \]
  where we have replaced $p$ by $\kappa$ since we $\rank(F) \leq \kappa$.
  The end of the proof is similar to the proof of Lemma \ref{lem:estimation}.
\end{proof}

We continue by reworking the approximation error.

\begin{lemma}
    When $\cL_p = \sum_{i\in[p]}\lambda_i f_i g_i^*$ is diagonalized by all the $\kappa=\binom{d+s}{d}$ polynomials of degree less or equal than $s$, $\norm{x} < M$ almost surely and $k_x(y) = ((1+x^\top y) / (1+M))^s$, then if the $\phi_i$ are almost surely linearly independent,
    \[
        \norm{(I-\Pi_G)\cL^{-1}} \leq \lambda_{\kappa+1}^{-1},\qquad\text{and}\qquad
        \norm{\Pi_G\cL^{-1}(I-\Pi_F)} = 0.
    \]
\end{lemma}
\begin{proof}
    In this case, the $\phi_i$'s span all the polynomials of degree less or equal than $s$, hence $(I-\Pi_F)\cL_p = 0$. 
\end{proof}

We continue by working out the terms $\norm{\Phi^{1/2}L^{-1}}$, $\norm{L^{-1}\Psi}$ and $\norm{L^{-1}}$ appearing in Theorem \ref{thm:stat-phi}.

\begin{lemma}
  When $\phi_i = \psi_i$ and $\norm{\Sigma^{-1/2}\phi}_{L^2(\rho)} \leq M_\infty$ with $M_\infty > 1/2$, and $\kappa = \rank(\Xi)$.
  For any $\delta \in (0, 1)$
  \[
	  p \geq 11 M_\infty^2\log(2\kappa/\delta),
  \]
  it holds with probability at least $1-\delta$,
  \[
	\max(\norm{\Psi^{1/2}L^{-1}}, \norm{L^{-1}\Phi^{1/2}}) \leq 2\norm{\Sigma^{-1/2} \cL^{-1}},
	\qquad\text{and}\qquad
	\norm{L^{-1}} \leq 4\norm{\Sigma^{-1/2} \cL^{-1} \Sigma^{-1/2}}.
  \]
\end{lemma}
\begin{proof}
  As detailed in the proof of Lemma \ref{lem:fake-constant},
  \begin{align*}
	\norm{L^{-1}\Psi^{1/2}}^2
	&= \norm{\hat\Xi^{-1/2}\cL^{-1}}^2
	= \norm{\cL^{-1}\hat\Xi^{-1}\cL^{-1}} 
	= \norm{\cL^{-1}\Xi^{-1/2}\Xi^{1/2}\hat\Xi^{-1}\Xi^{1/2}\Xi^{-1/2}\cL^{-1}} 
	\\&\leq \norm{\cL^{-1}\Xi^{-1/2}}^2 \norm{\Xi^{1/2}\hat\Xi^{-1}\Xi^{1/2}}
	= \norm{\cL^{-1}\Xi^{-1/2}}^2 \norm{(I - \Xi^{-1/2}(\Xi-\hat\Xi)\Xi^{-1/2})^{-1}}. 
  \end{align*}
  Using Bernstein inequality, we deduce that, if $M_\infty > 1/2$,
  \[
	\Pbb(\norm{\Xi^{-1/2}(\Xi - \hat \Xi)\Xi^{-1/2}} > 1/2) \leq
	2\kappa \exp(-\frac{3p}{32 M_\infty^2}).
  \]
  We invert the relation with
  \[
     \delta \leq 2\kappa\exp(-\frac{3p}{32M_\infty^2}),\qquad
	  p \geq \frac{32M_\infty^2}{3}\log(2\kappa/\delta).
  \]
  Similarly, under the same event
  \[
	\norm{L^{-1}} = \norm{\hat\Sigma^{-1/2}\cL^{-1}\hat\Sigma^{-1/2}}
	\leq\norm{\hat\Sigma^{-1/2}\Sigma^{1/2}}^2 \norm{\Sigma^{-1/2}\cL^{-1}\Sigma^{-1/2}}
	\leq 4 \norm{\Sigma^{-1/2}\cL^{-1}\Sigma^{-1/2}}.
  \]
  This explains the results of the lemma.
\end{proof}

\begin{proof}[Proof of Theorem \ref{thm:stat-L0}]
  Combining the previous lemmas, we have refinement of Theorems \ref{thm:stat-phi} and \ref{thm:stat} that when
  \[
	  n \geq 6\max(s^2\norm{\cL^{-1}}^{-2}, 1) \log(2\kappa/\delta),\qquad
	p \geq 11 \esssup_\phi\scap{\phi}{\Sigma^{-1}\phi}_{L^2(\rho)}\log(2\kappa/\delta),
  \]
  it holds with probability $1-2\delta$,
  \[
	  \big\| \cL^{-1} - G \hat L^{-1} F^* \big\|
	  \leq \lambda_{\kappa+1}^{-1} + 12 \|\Sigma^{-1/2} \cL^{-1}\|^2 \sqrt{\frac{8s^2}{3n}\log(\frac{2p}{\delta})}.
  \]
  To end the proof of the theorem, notice that when $\cL^{-1}$ is compact, the $\lambda_i^{-1}$ are summable hence they are bounded by $ci^{-\tau}$ for some constant $c$ and $\tau \geq 1$.
  Moreover, choosing
  \[
	  \kappa = \binom{s+d}{s} = \binom{d+s}{d} = \prod_{i\in[s]}\frac{(d+s-i+1)}{s-i+1)},
  \]
  because for all $k\in[s]$,
  \[
	  \frac{d+s}{s} \leq \frac{d+k}{k} \leq d,
  \]
  leads to  
  \[
	  (s/d)^d \leq (1+s/d)^d\leq \max((1+d/s)^s, (1+s/d)^d) \leq p \leq \min(d^s, s^d) \leq s^d.
  \]
  Hence our bound becomes
  \[
	  \big\| \cL^{-1} - G \hat L^{-1} F^* \big\|
	  \leq \lambda_{\kappa+1}^{-1} + 12 \|\Sigma^{-1/2} \cL^{-1}\|^2 \sqrt{\frac{8s^2}{3n}\log(\frac{2p}{\delta})}
	  \leq c d^{\tau d} s^{-\tau d} + \frac{20 s \norm{\Sigma^{-1/2}L^{-1}}^2}{n^{1/2}}\sqrt{\log\paren{\frac{p}{\delta}}}.
  \]
  Choosing $s = n^{1/2(\tau d+1)}$ leads to a bound $O(n^{-\tau d/2(\tau d+1)} \log(n))$, which holds as long as
  \[
	n \geq 6\max(n^{1/(\tau d+1)}\norm{\cL^{-2}}, 1)\paren{\frac{d}{2(\tau d+1)}\log(n) + \log(2/\delta)},
  \]
  and using that $\norm{K^{-1}\phi} \leq \mu_{\kappa+1}^{-1}\norm{\phi}$,
  \[
	n \geq p \geq 11 c n^{\tau d/2(\tau d+1)} M\paren{\frac{d}{2(\tau d+1)}\log(n) + \log(2/\delta)}.
  \]
  Both conditions can hold true for $p=n^{1/2}$ and $n > N$ for some $N\in\N$.
  The theorem in the main text considers the worst case where $\tau = 1$.
\end{proof}

\subsection{Variance of a gradient estimate through Jacobian vector products}
\label{app:var}

Consider the estimator
\[
    \norm{\nabla f(X)}^2 = \E_U[c\,\scap{\nabla f(X)}{U}^2], 
\]
for $U$ uniform on the sphere.
The proportionality constant is given by
\[
    c = \E_U[\scap{e_1}{U}^{2}]^{-1} = \E_{U}[U_1^2]^{-1} = \int_0^1 u_1^2 \sqrt{1-u_1^2} \vol(\bS^{d-2}) \diff u_1 = \frac{\pi}{16} \vol(\bS^{d-2}),
\]
which grows exponentially fast with respect to the input dimension $d$.
The estimator $G = c\,\scap{\nabla f(X)}{U}^2$ has a second moment, which assuming without restriction that $\norm{\nabla f(X)}^2 = 1$ is equal to
\begin{align*}
    \E[G^2] &= c^2 \E_U[\scap{e_1}{U}^{4}]^{-1} \geq \frac{c^2}{16}\Pbb(\abs{U_1} > 1/2) = \frac{c^2}{8\vol(\bS^{d-1})} \int_{1/2}^1 \sqrt{1-u_1^2} \vol(\bS^{d-2}) \diff u_1 
    \\&= \frac{c^2\vol(\bS^{d-2})}{8\vol(\bS^{d-1})}\paren{\frac{\pi}{6} - \frac{\sqrt{3}}{8}},
\end{align*}
while its mean is equal to one, hence its variance will grow exponentially fast as the dimension $d$ increases.
This explains the usefulness to restrict the loss \eqref{eq:tangent} to a few tangent directions, allowing to lower the variance of stochastic gradient descent and to accelerate its convergence \citep{Bubeck2015}.

%% file: appendix/experiments.tex
\section{Additional Experiments and Details}
\label{app:experiments}

\subsection{Graph-Laplacian Implementation}
\label{app:imp}

Many studies of graph-Laplacian are set in transductive settings, restricting $\X$ to a finite number of points \citep{Belkin2003,Zhu2003}.
In this study, we consider graph-Laplacian as a proxy to estimate $\cE_\cL$ empirically as per~\eqref{eq:graph-lap}.
As such, we can use Galerkin method with graph-Laplacian, only solving a eigenvalues problem associated with a $p\times p$ matrix, instead of the $n\times n$ Laplacian matrix, hence cutting cost from $O(n^3)$ flops to $O(p^3)$. A additional cost of the method lies in the building of the graph-Laplacian matrix, which requires $O(n^2p)$ flops, leading to a method scaling in $O(n^2 p + p^3 + np^2)$ (the last $O(np^2)$ being due to the building of $\Phi$).
Beside being statistically inferior and computationally more expensive, note that graph-Laplacian also introduces an extra hyperparameter, which is the function (and its scale) to compute the weight matrix $W$.

Another drawback of graph-Laplacian is that its convergence to the real Laplacian is known up to a constant \citep{Hein2007}, which rescales all eigenvalues.
To deal with this scaling constant, we assume $C = \sum_{i\in[k]} \lambda_i$ known for $\lambda_i$ the true eigenvalues of $\cL$ and $k=25$ used to evaluate eigenvalues retrieval as per Figure~\ref{fig:eigen}, and we scale the graph Laplacian to ensure $\sum_{i\in[k]}\hat\lambda_i = C$.
This can only improve the performance of graph-Laplacian, and only reinforce our findings on the superiority of Galerkin method, for which we do not employ this trick.

\begin{table}[t]
    \centering
    \begin{tabular}{|c|c|c|}
    \hline
       & Graph-Laplacian & Kernel-Laplacian \\
    \hline
       Time complexity & $n^2 d + p^2 n$ & $pnd + p^2 n$ \\
       Memory complexity & $n^2$ & $pn$ \\
    \hline
    \end{tabular}
	\vspace{.5em}
    \caption{{\bf Computational complexity of graph-Laplacian and kernel-Laplacian}. Not only kernel-Laplacian does not suffer from the curse of dimension, which contrasts with graph-Laplacian, but it does so without requiring extra computations (recall that $p \leq n$ is taken as a small integer).}
    \label{tab:my_label}
\end{table}

\subsection{Additional Figures}
\begin{figure}[t]
    \centering
	\includegraphics{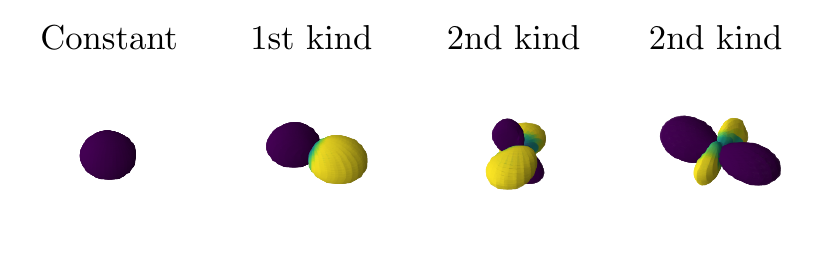}
	\includegraphics{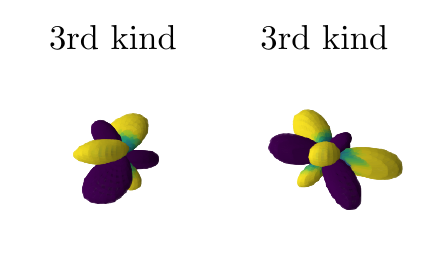}
    \vspace{-1em}
	\caption{
	  Cherry-picked learned spherical harmonics with a neural network.
	  The network is a multi-layer perceptron with 200, 200, 2000, 200 hidden neurons in the four hidden layers, and $m=16$ outputs optimized over 5000 batches of size 1000 with the contrastive version of the orthogonal regularizer.
	  The optimizer is stochastic gradient descent with momentum ($m=1/2$) initialized with a learning rate $\gamma = 10^{-3}$, with a scheduler to decrease the learning rate after one third and two third of the learning by a factor $1/3$.
	  Principal component analysis was used to disentangle the learned representation and retrieve the different learned eigenspaces and eigenfunctions.
	}
    \label{fig:sph-nn}
    \vspace{-1em}
\end{figure}

\begin{figure}[t]
    \centering
	\includegraphics[width=.48\linewidth]{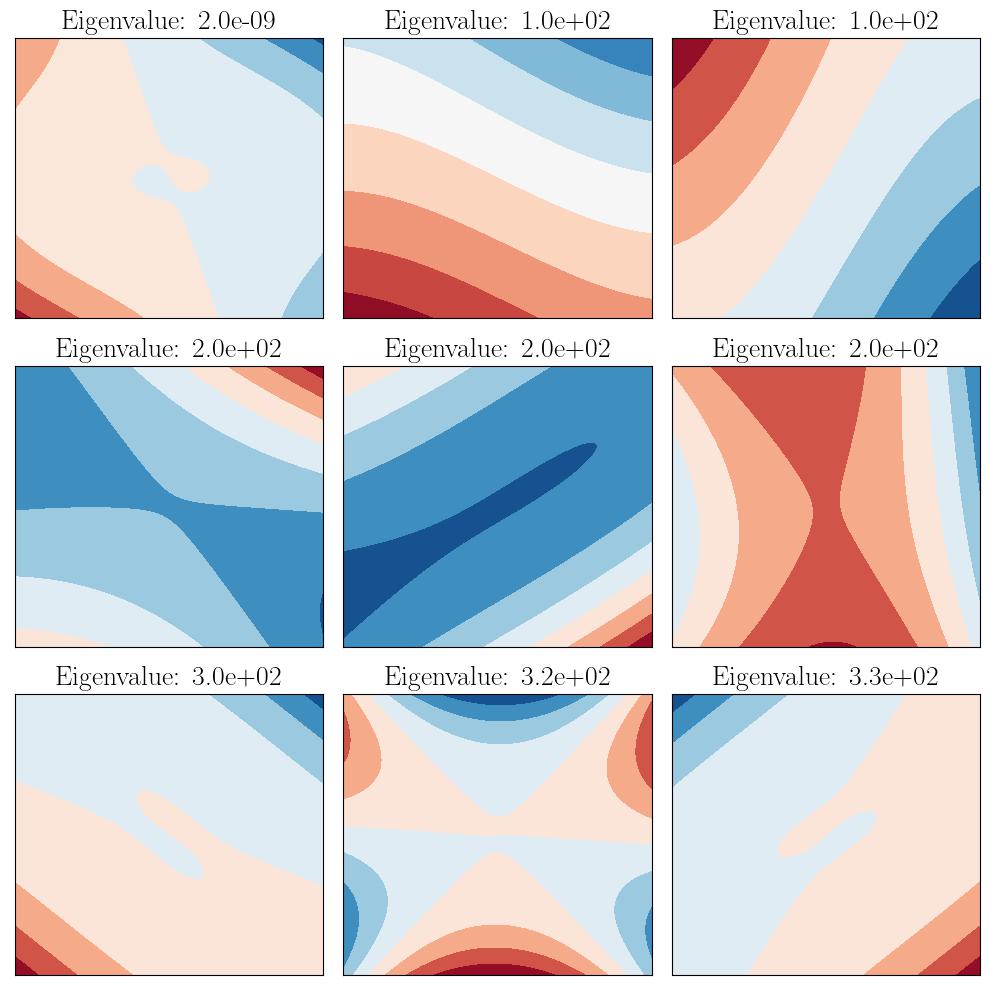}
	\vline{}
	\includegraphics[width=.48\linewidth]{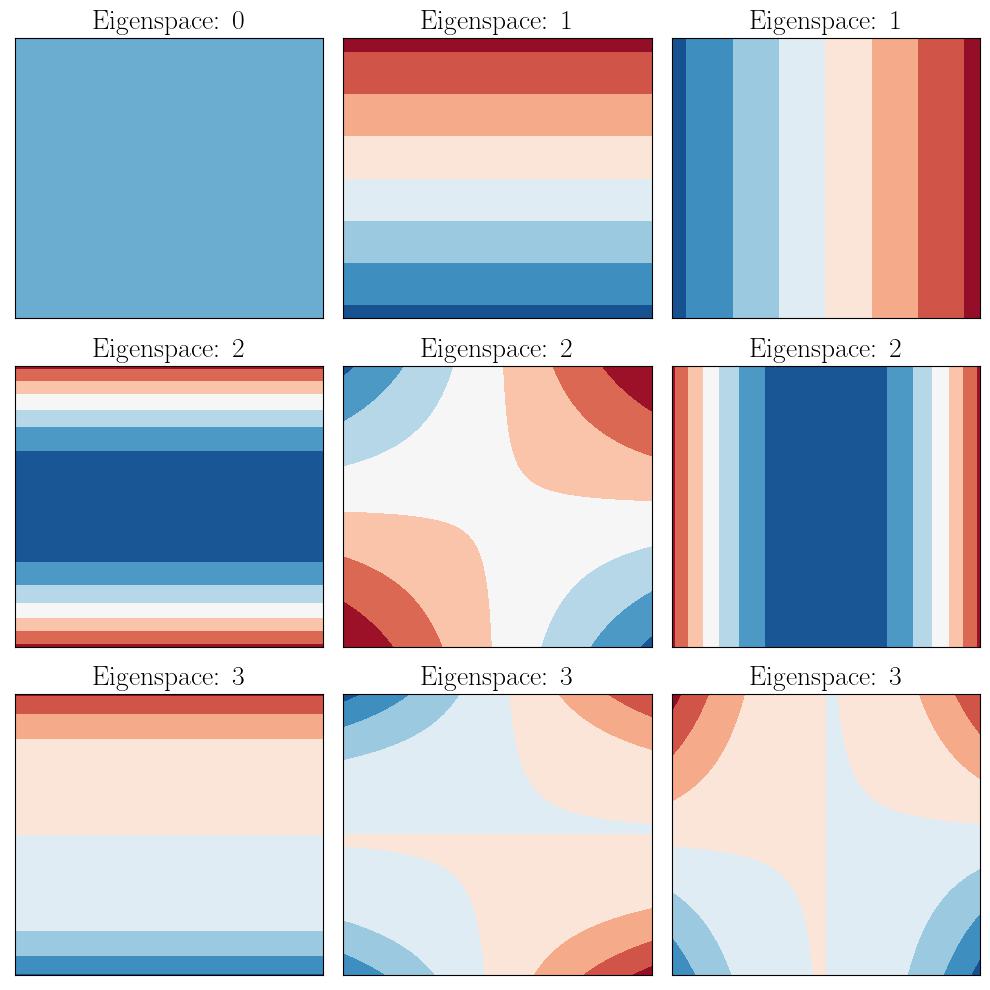}
	\caption{
	  (Left) Level lines of the learned hermite polynomials in $2d$ thanks to the operator $\cL$ with $\X = \R^2$ and $\rho = \cN(0, I)$, with polynomial kernel of degree 3.
	  Compared to the ground truth on the right, notice how the learned polynomials are ``random'' basis of the different eigenspace of $\cL$.
	  (Right) Canonical Hermite polynomials, acting as a ground truth for the left part.
	  The title indicates the eigenspace number that the eigenfunctions belong to.
	  Any orthogonal transformation $U(f_i)$ for $(f_i)$ the $k$ eigenfunctions of the $k$-th eigenspace, and $U$ an orthogonal matrix in $\R^{k\times k}$ is a valid basis of eigenfunctions of this eigenspace, which is what is found in practice.
	}
    \label{fig:hermite-2d}
    \vspace{-1em}
\end{figure}

To support empirically the claim made in Section~\ref{sec:nn}, Figure~\ref{fig:sph-nn} illustrates the learning of spherical harmonics with a neural network.
Finally, Figure~\ref{fig:hermite-2d} shows Hermite polynomials in two dimensions, corresponding to the eigenfunctions of $\cL_0$ with $\rho = \cN(0, I)$, which could have served as a different basis to study our method.
However, in high-dimension, $\cN(0, I)$ tends to concentrate on the sphere and we do not expect many different behaviors in comparison to our study with spherical harmonics.

\clearpage
\subsection{Empirical Observations}
While Figure~\ref{fig:eigen} illustrates our take-home messages, i.e., ``Galerkin beats graph-Laplacian, and its does not cost much in terms of implementation (scaling linearly with respect to $n$ and being almost indifferent to the dimension)'', much more could be said when digging into the one hundred runs with all the different hyperparameters.

\paragraph{Parameter grid.}
We consider the following values for the different experimental setups with spherical harmonics.
\begin{itemize}
	\item $n$: ten values equally spaced in log space between 100 and 10000.
	\item $d \in \brace{3, 5, 7, \ldots, 19}$
	\item $p$ five values equally spaced in log space between 30 and 1000.
\end{itemize}
We try three kernels for the Galerkin functions, together with five hyperparameters for each.
\begin{itemize}
  \item The polynomial kernel $k_x(y) = (1+x^\top y)^s$ with hyperparameter $s \in \brace{2, 3, 4, 5, 6}$.
  \item The exponential kernel $k_x(y) = \exp(-\norm{x-y}/\sigma)$ with hyperparameters $\sigma \in \brace{.1, 1, 10, 100, 1000}$.
  \item The Gaussian kernel $k_x(y) = \exp(-\norm{x-y}^2/2\sigma^2)$ with hyperparameter $\sigma \in \brace{.01, .1, 1, 10, 100}$
\end{itemize}
For the Graph-Laplacian, we equally consider six different options for the weighting scheme in~\eqref{eq:graph-lap}.
\begin{itemize}
	\item Either $w_{ij} = k_{x_i}(x_j)$ with the same kernel as the Galerkin functions.
	\item Or $w_{ij} = \exp(-\norm{x_i-x_j)}^2 / 2\sigma^2)$ with hyperparameters $\sigma \in \brace{.01, .1, 1, 10, 100}$.
\end{itemize}
The result of Figure~\ref{fig:eigen} were taken as the best over both $p$ and each kernel and hyperparameters, leading to the best pick out of $5\times 3\times 5 = 75$ options for Galerkin, and out of $6 \times 75=450$ for graph-Laplacian.
We ran one hundred trials for each configuration, and used Slurm to parallelize runs on a cluster of CPUs, together with the {\tt SeedSequence} generator of Numpy to ensure a minimum correlation between pseudo-random runs \citep{Harris2020}.

\begin{figure}[t]
    \centering
        \includegraphics{images/legend_varying_d.pdf}\\
	\includegraphics{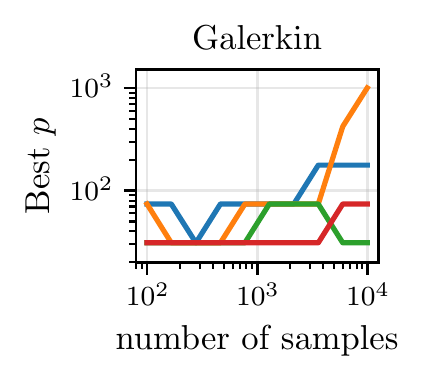}
	\includegraphics{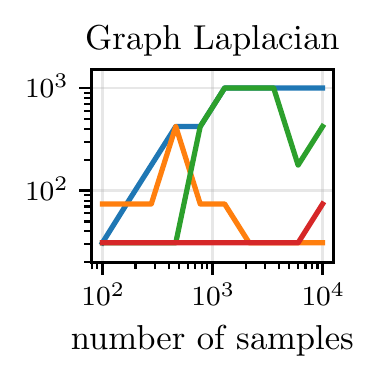}
	\includegraphics{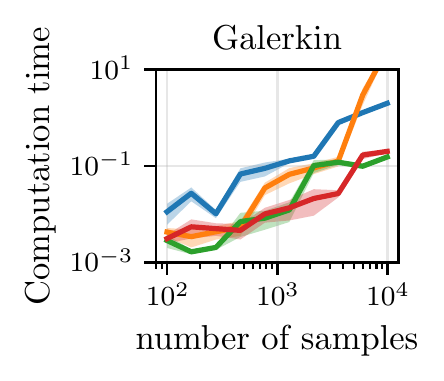}
	\includegraphics{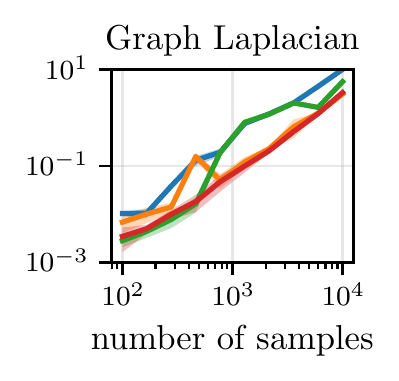}
    \vspace{-1em}
	\caption{
        (Left) Value of $p$ achieving the best results reported on Figure~\ref{fig:eigen}.
        (Right) Computation time corresponding to the best results.
	}
    \label{fig:best-time}
\end{figure}

\begin{figure}[t]
    \centering
        \includegraphics{images/legend_varying_d.pdf}\\
	\includegraphics{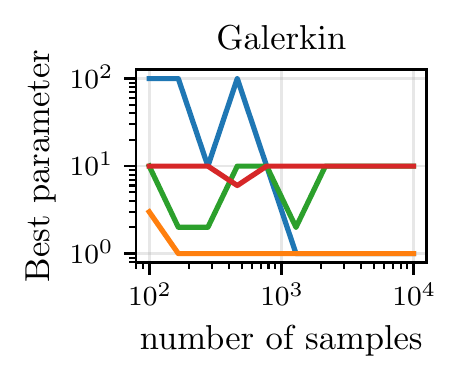}
	\includegraphics{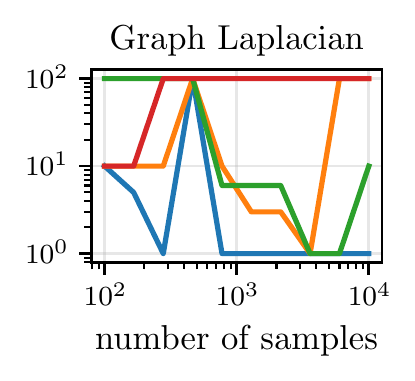}
	\includegraphics{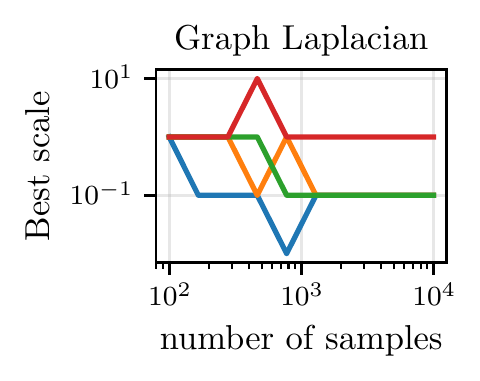}
    \vspace{-1em}
	\caption{
        (Left) Best parameters for Galerkin functions, they are strictly inside our grid of parameters (for graph-Laplacian, the best results are obtained with the exponential kernel for which our grid stops at $10^3$).
        (Right) Best parameters for the weights $w_{ij}$ in~\eqref{eq:graph-lap}. Once again, the best values are obtained strictly inside the grid.
	}
    \label{fig:best-hyper}
\end{figure}

\paragraph{Best results.}
First of all, we start by looking at the best results.
The left of Figure~\ref{fig:best-time} shows that the best values of $p$ are not always the biggest ones.
This is understandable, a bigger number of ``Galerkin'' functions leads to a bigger risk of overfitting, in particular in high-dimensions, which explains why the red curves are below the other ones.
The right of Figure~\ref{fig:best-time} showcases the corresponding computation time, they are highly correlated with the value of $p$, and they are of similar order for graph-Laplacian and the Galerkin method.
Finally, Figure~\ref{fig:best-hyper} makes sure that the best hyperparameter values were not obtained for extreme values of our parameter grid, removing confounders due to bad hyperparameters calibration.

\begin{figure}[t]
    \centering
	\includegraphics{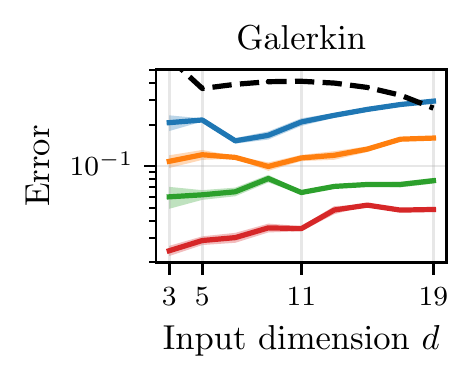}
	\includegraphics{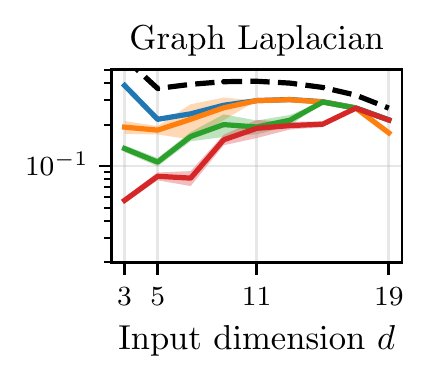}
        \hspace{1em}
        \includegraphics{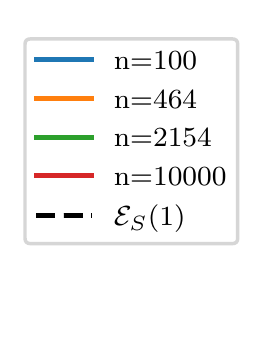}
    \vspace{-1em}
	\caption{
        {\em Effect of the dimension.} Scaling of the error~\eqref{eq:sur} as a function of the input dimension $d$ for the best set of hyperparameters for both the Galerkin method and graph-Laplacian.
        Note that the problem changes as the dimension augments, and that we are rescaling the error so that $\cE_S(0) = 1$ regardless of the dimension.
        The graph-Laplacian seems to reach a default error as $d$ increases which should be put in comparison with the error reached by $\hat\lambda_i = 1 / \sum_{i\in[k]}\lambda_i$ plotted in black (recall that we artificially rescaled the graph-Laplacian to ensure $\sum_{i\in[k]}\hat\lambda_i = \sum_{i\in[k]}\lambda_i$).
	}
    \label{fig:scale-d}
\end{figure}

\begin{figure}[t]
    \centering
        \includegraphics{images/legend_varying_d.pdf}\\
	\includegraphics{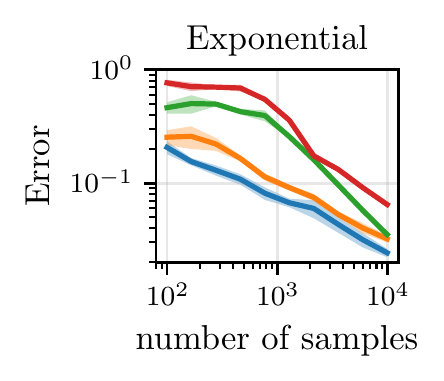}
	\includegraphics{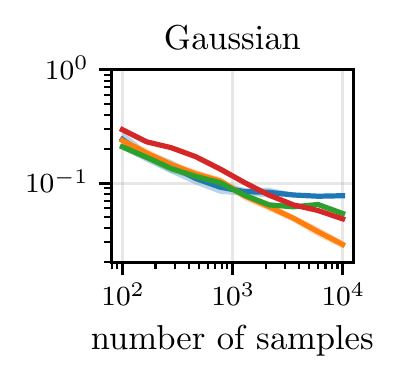}
	\includegraphics{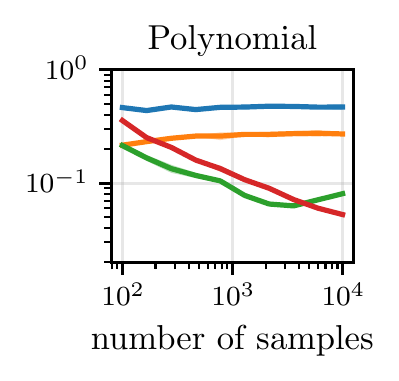}\\
	\includegraphics{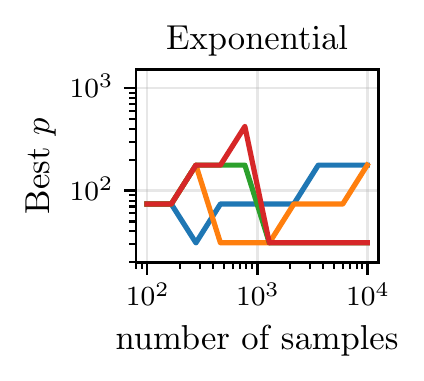}
	\includegraphics{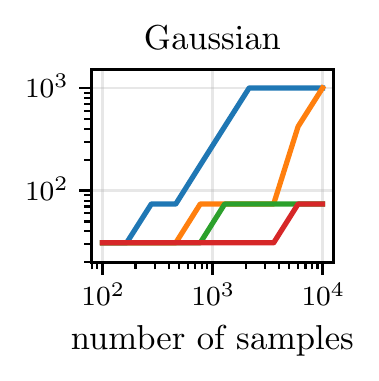}
	\includegraphics{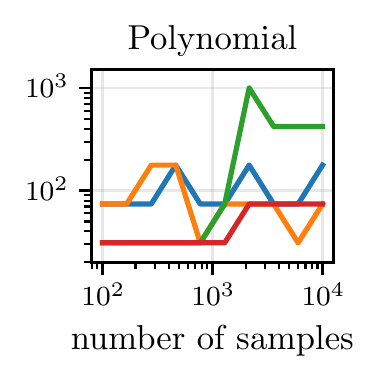}\\
	\includegraphics{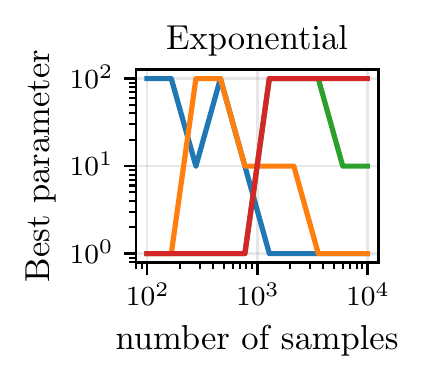}
	\includegraphics{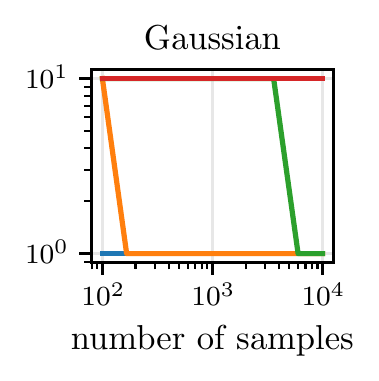}
	\includegraphics{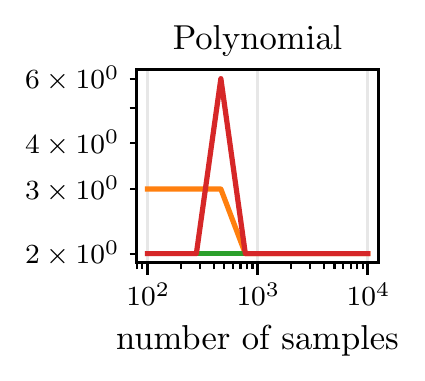}
    \vspace{-1em}
	\caption{
        Influence on the kernel for Galerkin method.
	}
    \label{fig:kernel}
\end{figure}

\paragraph{Effect of the different parameters at play.}
Figure~\ref{fig:scale-d} explores the effect of the dimension on our estimator quality.
Figure~\ref{fig:kernel} explores the effect of the kernel used for the Galerkin method. 
We notice that although the eigenfunctions are polynomials, the polynomial kernel does not necessarily lead to the best result, this is due to the fact that there are many polynomials in dimension $d$, and the polynomial kernel does favor the learning of simple polynomials over more complex ones.
We again notice the regularizing effect of not choosing the biggest $p$ possible (e.g., choosing $p=1000$ as soon as $n\geq 1291$ according to our grid values).

\clearpage
\subsection{Hermite Regression}
\label{app:hermite}

Consider the Hermite interpolation problem, where for $(x_i)_{i\in[n]}$ $n$ data point in $\X=\R^d$, $(v_i)$ $n$ scalar values, and $(t_i)$ $n$ vectors in $\R^d$, we try to solve
\[
    \argmin_{f\in\cF}\sum_{k\in[n]} \norm{f(x_k) - v_k}^2 + \norm{\nabla f(x_k) - t_k}^2,
\]
for some space of functions $\cF$.
In our case, we consider
\[
    \cF = \Span\brace{\phi_i \midvert i\in[p]},
\]
which will not ensure interpolation when $p < n(d+1)$, hence our denomination of ``Hermite regression''.
With $\alpha\in\R^p$ and $f = F\alpha = \sum_{i\in[n]}\alpha_i \phi_i$, this leads to 
\begin{align*}
    &\argmin_{\alpha\in\R^p}\sum_{k\in[n]}\big\|\sum_{i\in[p]}\alpha_i \phi_i(x_k) - v_k\big\|^2 + \big\|\sum_{i\in[p]}\alpha_i\nabla \phi_i(x_k) - t_k\big\|^2
    \\&=\argmin_{\alpha\in\R^p}\sum_{k\in[n]}\sum_{i,j\in[p]}\alpha_i \alpha_j\paren{\phi_i(x_k)\phi_j(x_k) + \sum_{l\in[d]}\partial_l \phi_i(x_k)\partial_l\phi_j(x_k)} -2\sum_{i\in[p]}\alpha_i \paren{\phi_i(x_k) v_k + \sum_{l\in[d]}\partial_l\phi_i(x_k) t_{kl}}
    \\&=\argmin_{\alpha\in\R^p} \alpha^\top A \alpha - 2b^\top \alpha = A^{-1}b,
\end{align*}
where $A\in\R^{p\times p}$ and $b\in \R^p$ are defined as
\[
    A_ij = \sum_{k\in[n]}\phi_i(x_k)\phi_j(x_k) + \scap{\nabla\phi_i(x_k)}{\nabla\phi_j(x_k)}, \qquad b_i = \sum_{k\in[n]}\phi_i(x_k) v_k + \scap{\nabla\phi_i(x_k)}{t_k}.
\]
A naive implementation leads to $O(np^2d + npd)$ flops and $O(p^2)$ bits to build those matrices, and $O(p^3)$ to solve $\alpha = A^{-1}b$.
In the case where $\phi_i = k_{x_i}$ with $k$ a dot-product or an invariant kernel, Proposition~\ref{thm:imp} shows how to reduce the time complexity to $O(np^2 + npd)$ flops at the expense of a memory complexity in $O(p^2 + np)$ bits.
Note that similarly to Proposition~\ref{thm:imp} the term $b$ can also be written in a simple form for structured kernels, although this does not reduce the overall complexity of building $b$.

\begin{proposition}
\label{prop:imp}
  Assume that $\X$ is endowed with a scalar product.
  Given a kernel $k_x(y) = q(\norm{x - y})$ defined from $q:\R\to\R$, for $x, y, t\in \X$, it holds
  \begin{equation}
     \scap{\nabla k_y(x)}{t} = \frac{q'(\norm{x-y})}{\norm{x-y}}\,(x-y)^\top t.
  \end{equation}
  Similarly for dot-product kernel $k_x(y) = q(x^\top y)$, 
  \begin{equation}
     \scap{\nabla k_y(x)}{t} = q'(x^\top y)\, y^\top t.
  \end{equation}
\end{proposition}
\begin{proof}
  Once again, the proof follows from the application of the chain rule.
\end{proof}

Propositions~\ref{thm:imp} and~\ref{prop:imp} explain our Hermite regression algorithms with dot-product kernel, Algorithm~\ref{alg:hermite-dot}, and translation-invariant kernel, Algorithm~\ref{alg:hermite}.

\begin{algorithm}[ht]
    \KwData{Data $(x_i) \in \R^{n\times p}$, $V = (v_i)\in\R^n$, $(t_i)\in\R^{n\times p}$, kernel $k_x(y) = q(x^\top y)$.}
    Compute $X = (x_i^\top x_j) \in \R^{p\times n}$ with $p \leq n$;\\ 
    Compute $\Psi = q(X) q(X)^\top \in \R^{p\times p}$ elementwise;\\
    Compute $L = (q'(X) q'(X)^\top)$;\\
    Update $L_{ij} \leftarrow X_{ij} L_{ij}$ for all $i, j\in[p]$;\\
    Set $A = L + \Psi$;\\
    Compute $T = (x_i^\top t_j)\in\R^{p\times n}$;\\
    Update $T \leftarrow q'(X) \odot T$ with elementwise (Hadamard) product;\\
    Compute $b = q(X) V + T 1_n \in \R^p$ where $1_n = (1)_{i\in[n]} \in \R^n$;\\
    Solve $\alpha = A^{-1} b$;\\
    Set $\hat f(x) := \sum_{i\in[p]} \alpha_{i} k_{x_i}(x)$.\\
    \KwResult{Hermite estimator $\hat f$}
    \caption{Hermite regression with dot-product kernel}
    \label{alg:hermite-dot}
\end{algorithm}

\begin{algorithm}[ht]
    \KwData{Data $(x_i) \in \R^{n\times p}$, $V = (v_i)\in\R^n$, $(t_i)\in\R^{n\times p}$, kernel $k_x(y) = q(\norm{x - y})$.}
    Compute $X = (x_i^\top x_j), \in \R^{p\times n}$, $D = (x_i^\top x_i) \in \R^n$;\\
    Deduce $N = (\norm{x_i - x_j}) \in \R^{p\times n}$ and $T = q'(N) / N$;\\
    Initialize $L = 0 \in \R^{p\times p}$;
    $\Psi$ = $q(N) q(N)^\top \in \R^{p\times p}$;\\
    \For{$k \in [n]$}{
    Set $\gamma^{(k)}_{ij} := (D_k - X_{ik} - X_{jk} + X_{ij})$;\\
    Update $L_{ij} \leftarrow L_{ij} + \gamma_{ij}^{(k)} T_{ik} T_{jk}$;\\
    }
    Set $A = L + \Psi$;\\
    Compute $T = -(x_i^\top t_j)\in\R^{p\times n}$;\\
    Update $T_{ij} \leftarrow T_{ij} + x_j^\top t_j$ for all $i\in[p]$, $j\in[n]$;\\
    Update $T \leftarrow T \odot (q'(N) / N)$ with elementwise (Hadamard) product;\\
    Compute $b = q(N) V + T 1_n \in \R^p$ where $1_n = (1)_{i\in[n]} \in \R^n$;\\
    Solve $\alpha = A^{-1} b$;\\
    Set $\hat f(x) := \sum_{i\in[p]} \alpha_{i} k_{x_i}(x)$.\\
    \KwResult{Hermite estimator $\hat f$}
    \caption{Hermite regression with distance kernel}
    \label{alg:hermite}
\end{algorithm}